%% file: main.tex
\documentclass[runningheads]{llncs}

\usepackage[T1]{fontenc}
\usepackage{graphicx}
\usepackage{booktabs}
\usepackage{amsmath}
\usepackage{amssymb}
\usepackage[table]{xcolor}
\usepackage{multirow}
\usepackage{array}
\usepackage{subcaption}
\usepackage{enumitem}
\usepackage{placeins}
\usepackage{float}
\usepackage{makecell}
\usepackage{xspace}

\input{_macros}
\usepackage{hyperref}
\hypersetup{colorlinks=true, linkcolor=blue, citecolor=blue, urlcolor=blue}

\usepackage[capitalize]{cleveref}
\crefname{section}{Sec.}{Secs.}
\crefname{table}{Table}{Tables}
\crefname{figure}{Fig.}{Figs.}
\crefname{equation}{Eq.}{Eqs.}

\begin{document}

\title{BOCCHI: A More Realistic and Challenging Benchmark for Local Motion Blur Detection with MSDCT-UNet}

\titlerunning{BOCCHI and MSDCT-UNet for Local Motion Blur Detection}

\author{Kuan-Lin Chen\inst{1}\orcidID{0000-0002-0491-5713} \and
        Yuan-Kang Lee\inst{2}\orcidID{0009-0003-0468-3734} \and
        Cheng-Yuan Chiang\inst{1}\orcidID{0009-0001-0130-444X} \and
        Jian-Jiun Ding\inst{1}\orcidID{0000-0003-4510-2273}}
\authorrunning{K.-L. Chen et al.}
\institute{Graduate Institute of Communication Engineering,\\
National Taiwan University, Taipei, Taiwan\\
\email{r13942067@ntu.edu.tw, p14942a02@ntu.edu.tw, jjding@ntu.edu.tw}
\and
MediaTek Inc., Hsinchu, Taiwan\\
\email{r12942062@ntu.edu.tw}}

\maketitle

\begin{figure}[t]
    \centering
    \includegraphics[width=\linewidth]{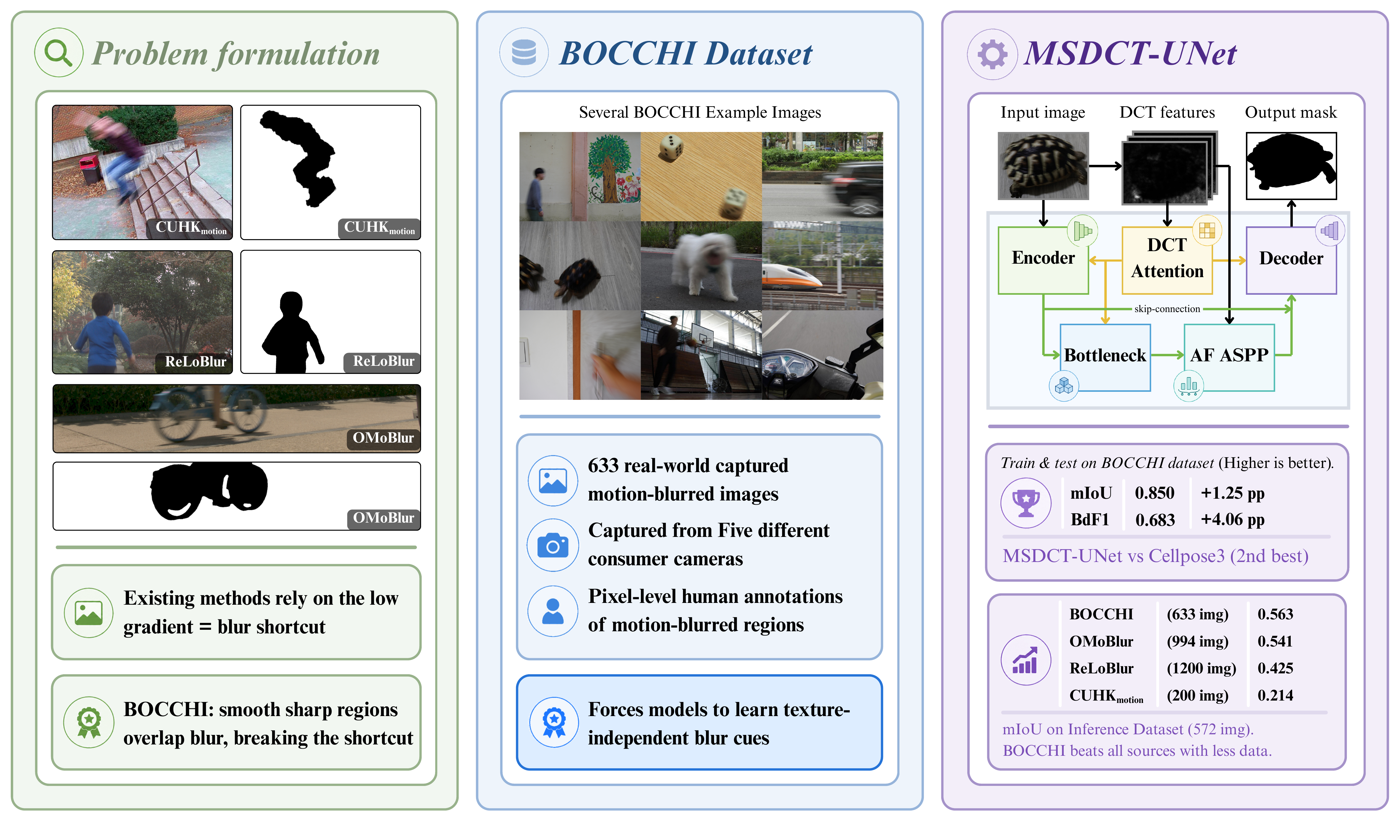}
    \caption{\textbf{Overview of our contributions.} Existing benchmarks can permit gradient-based shortcuts when blur and sharp regions are easily separable. \textbf{BOCCHI} addresses this with 633 real-captured, pixel-annotated images whose sharp regions cover both textured and smooth surfaces, creating strong gradient overlap with blurred objects. Built on this benchmark, \textbf{MSDCT-UNet} injects multi-scale DCT priors and achieves the best in-domain performance on BOCCHI, while BOCCHI-trained models show the strongest cross-dataset transfer with fewer training samples.}
    \label{fig:teaser}
\end{figure}

\input{00_abstract}
\input{01_intro}
\input{02_related}
\input{03_dataset}
\input{04_method}
\input{05_experiment}
\input{10_conclusion}

\clearpage
\begin{center}
{\Large \bfseries\boldmath Supplementary Material for: BOCCHI: A More Realistic and Challenging Benchmark for Local Motion Blur Detection with MSDCT-UNet\par}
\vskip .8cm
{\lineskip .5em\noindent
Kuan-Lin Chen\inst{1}\orcidID{0000-0002-0491-5713},
Yuan-Kang Lee\inst{2}\orcidID{0009-0003-0468-3734},
Cheng-Yuan Chiang\inst{1}\orcidID{0009-0001-0130-444X}, and
Jian-Jiun Ding\inst{1}\orcidID{0000-0003-4510-2273}
\par\vskip.35cm}
{\small
\inst{1}Graduate Institute of Communication Engineering,
National Taiwan University, Taipei, Taiwan\\
\email{r13942067@ntu.edu.tw, p14942a02@ntu.edu.tw, jjding@ntu.edu.tw}\\
\inst{2}MediaTek Inc., Hsinchu, Taiwan\\
\email{r12942062@ntu.edu.tw}\par}
\end{center}
\vskip .5cm

\input{_supp_body}

\clearpage
\bibliographystyle{splncs04}
\bibliography{references}

\end{document}

%% file: _macros.tex
\newcolumntype{C}{>{\centering\arraybackslash}p{1.05cm}}

\definecolor{lightred}{RGB}{253,191,191}
\definecolor{lightorange}{RGB}{255,223,191}
\definecolor{lightyellow}{RGB}{254,240,198}

%% file: 00_abstract.tex
\begin{abstract}

Local motion blur detection requires pixel-level localization of blurred regions. Existing benchmarks let models rely on gradient shortcuts that fail to transfer. We introduce \textbf{BOCCHI} (Blurred Objects Captured across Cameras with Human-annotated Imagery), a real-captured benchmark whose sharp regions overlap the blur gradient distribution and defeat these shortcuts, and propose \textbf{MSDCT-UNet} (Multi-Scale Discrete Cosine Transform UNet), a frequency-aware encoder-decoder injecting multi-scale DCT priors through DCT Attention and FiLM. MSDCT-UNet ranks first in in-domain mIoU and boundary localization on BOCCHI, and BOCCHI-trained models outperform every other training source on cross-dataset transfer with only $633$ training images. Our project page is available at \url{https://brianchen1120.github.io/project/bocchi/}.

\keywords{Motion blur detection \and Local motion blur dataset \and DCT attention \and Frequency-domain learning \and Semantic segmentation}

\end{abstract}

%% file: 01_intro.tex
\section{Introduction}
\label{sec:intro}


When an object moves during camera exposure, the resulting image exhibits spatially non-uniform local motion blur, where only moving objects appear blurred while the background remains sharp. Localizing such partial blur at pixel-level precision is challenging, and an accurate blur mask supports applications such as selective deblurring~\cite{zamir2022restormer,li2024lmdvit}, confidence calibration in object detection under motion, and restoration preprocessing. We formulate this problem as binary semantic segmentation.


Existing methods (surveyed in \cref{sec:related}) mostly rely on hand-crafted frequency descriptors~\cite{tong2004blur,shi2014discriminative,golestaneh2017hifst,ma2018deepblur} or spatial-only learning~\cite{sun2015learning,gong2017motion,zamir2022restormer,tsai2022stripformer,hu2022decomposed,lumentut2020human}, neither adapting well to diverse motion-blur scenes. We argue this arises partly from benchmark limitations: CUHK$_{\text{motion}}$~\cite{shi2014discriminative} has only around 200 images, while ReLoBlur~\cite{li2023reloblur} concentrates on street scenes where the blurred regions are typically small and sit against textured backgrounds. Models trained on these benchmarks may learn shortcuts, e.g., treating low-frequency regions as blurred, without modeling the underlying optics of motion blur. To probe this harder regime, we introduce a real-captured benchmark whose sharp regions overlap the blur-region gradient distribution. As quantified in \cref{sec:dataset}, this overlap is larger on BOCCHI than on any other training source, breaking the low-gradient-equals-blur shortcut and requiring models to reason about frequency cues rather than spatial gradient alone. The main contributions of this paper are:


\begin{enumerate}
    \item \textbf{BOCCHI benchmark.} We introduce a real-captured dataset of 633 camera images exposing a harder local blur regime in which sharp regions span both textured backgrounds and smooth surfaces, creating gradient-distribution overlap with the blur region that is largely absent from prior datasets.

    \item \textbf{MSDCT-UNet.} To our knowledge, the first segmentation network that injects multi-scale DCT priors at every encoder-decoder stage via a multi-head DCT Attention, FiLM modulation, and an AFASPP bottleneck, explicitly disentangling blur evidence from spatial appearance.

    \item \textbf{Empirical study.} Against 12 baselines: \emph{(i)} MSDCT-UNet ranks first in-domain on BOCCHI ($+1.2$ pp mIoU); \emph{(ii)} BOCCHI-trained models outperform all three other training sources (ReLoBlur, CUHK$_{\text{motion}}$, OMoBlur) on cross-dataset mIoU/Dice/Recall averages with only $633$ training images, a dataset-level data efficiency.
\end{enumerate}

%% file: 02_related.tex
\section{Related Work}
\label{sec:related}

\paragraph{Blur Detection.}
Traditional methods detect blur via hand-crafted frequency features: Haar wavelet energy decay~\cite{tong2004blur}, discriminative gradient/spectral/saturation features ~\cite{ji2008motion,shi2014discriminative,lee2024efficient}, local power spectrum slopes~\cite{yang2005motion,liu2008noise}, and multi-scale DCT coefficients on gradient magnitudes (HiFST~\cite{golestaneh2017hifst}), which directly inspires our DCT branch design.
Deep learning methods recast blur analysis as kernel prediction from patches~\cite{sun2015learning}, per-pixel motion flow estimation~\cite{gong2017motion}, or defocus detection via depth distillation~\cite{cun2020defocus}. Reinforcement learning has been applied to various vision tasks recently~\cite{lee2026rl,lee2024learning,zhang2023image}, including blur-region detection, where AlignSAM uses an RL agent to select point prompts for segmentation~\cite{huang2024alignsam}. Also, deblurring architectures such as Restormer~\cite{zamir2022restormer} and Stripformer~\cite{tsai2022stripformer} extract powerful blur-aware features but output restored images rather than explicit blur masks.
\emph{Unlike} these methods relying on \emph{fixed} hand-crafted features or \emph{implicit} spatial-domain learning, we explicitly extract multi-scale DCT features and learn \emph{adaptive} frequency attention weights, combining classical frequency analysis with deep representation power.

\paragraph{Dense Prediction Architectures.}
We formulate local blur detection as binary segmentation, building on encoder-decoder designs with skip connections~\cite{ronneberger2015unet} and multi-scale context aggregation via ASPP~\cite{chen2018deeplabv3plus} or pyramid pooling~\cite{zhao2017pspnet}.
Lightweight architectures pursue real-time inference through bilateral branches~\cite{yu2021bisenetv2}, short-term dense concatenation~\cite{fan2021stdc}, dual-resolution feature exchange~\cite{hong2022ddrnet}, and depthwise separable convolutions~\cite{poudel2019fastscnn}; heavier models maintain multi-resolution representations~\cite{wang2020hrnet} or adopt modernized ConvNet blocks with large kernels and inverted bottlenecks~\cite{liu2022convnext}.
Multi-scale convolutional attention~\cite{guo2022segnext} and activation-free restoration baselines~\cite{chen2022nafnet,yang2020densedual} further push accuracy--efficiency trade-offs.
\emph{However}, these architectures process \emph{only} spatial-domain features and lack inductive biases for blur-specific frequency characteristics; our MSDCT-UNet augments a ConvNeXt-style backbone with a \emph{parallel DCT attention branch} at every encoder--decoder stage, injecting frequency-domain priors critical for distinguishing blurred from sharp regions.

\paragraph{Frequency-Domain Learning.}
Recent works integrate frequency representations into deep networks: block-wise DCT coefficients as CNN inputs~\cite{xu2020frequency}, multi-spectral DCT bases generalizing SE channel attention~\cite{qin2021fcanet,hu2018senet}, focal frequency loss emphasizing hard-to-reconstruct spectral components~\cite{jiang2021focal}, and adaptive frequency filters as efficient global token mixers~\cite{huang2023freqnet,yang2022affcam}.
For heterogeneous feature fusion, FiLM~\cite{perez2018film} introduces feature-wise affine modulation, and CBAM~\cite{woo2018cbam,liu2022ffca} provides sequential channel--spatial recalibration.
\emph{In contrast to} methods using \emph{fixed} DCT bases~\cite{qin2021fcanet}, \emph{auxiliary} frequency losses~\cite{jiang2021focal}, or generic spatial mixers~\cite{huang2023freqnet}, our DCT Attention computes \emph{position-dependent, multi-head} attention over 57 frequency channels with SE-gating and temperature-scaled softmax, fused via FiLM conditioning, providing a \emph{task-specific} inductive bias that explicitly models the high-frequency attenuation characteristic of motion blur.

%% file: 03_dataset.tex
\section{BOCCHI Dataset}
\label{sec:dataset}

\subsection{Motivation}

Existing benchmarks are insufficient for real-world local motion-blur detection: CUHK$_{\text{motion}}$~\cite{shi2014discriminative} is small, ReLoBlur~\cite{li2023reloblur} is scene- and blur-ratio-biased, and synthetic GoPro-style data~\cite{nah2017gopro} does not reflect real PSF-driven blur. A contemporaneous work, OMoBlur~\cite{yu2026omoblur}, releases physically grounded local-motion-blur image pairs derived from high-speed video frame averaging; its optical-flow-derived masks, however, are soft restoration supervision targets rather than manual detection annotations. We therefore introduce \textbf{BOCCHI}, a benchmark of $633$ real-captured images across cameras with pixel-level human annotations; collection details are given in \cref{subsec:collection}.

\subsection{Collection and Annotation}
\label{subsec:collection}

\textbf{Collection.} BOCCHI was captured in-the-wild using five consumer-grade cameras spanning five different manufacturers: a SONY $\alpha$6400, a Nikon D60, an Olympus E-M5, a Canon EOS R8, and a Fujifilm X-T30. Images were shot at each camera's native resolution, then center-cropped to a 3:2 aspect ratio, yielding $1080 \times 720$ (landscape) or $720 \times 1080$ (portrait) images; portrait images are rotated to landscape during preprocessing so that all network inputs share a uniform $1080 \times 720$ resolution. The scenes span diverse indoor and outdoor environments, including streets, parks, shopping malls, sports fields, and campus walkways. The dynamic subjects include pedestrians, vehicles, animals, hand gestures, and sports equipment.

\begin{figure}[t]
    \centering
    \includegraphics[width=\linewidth]{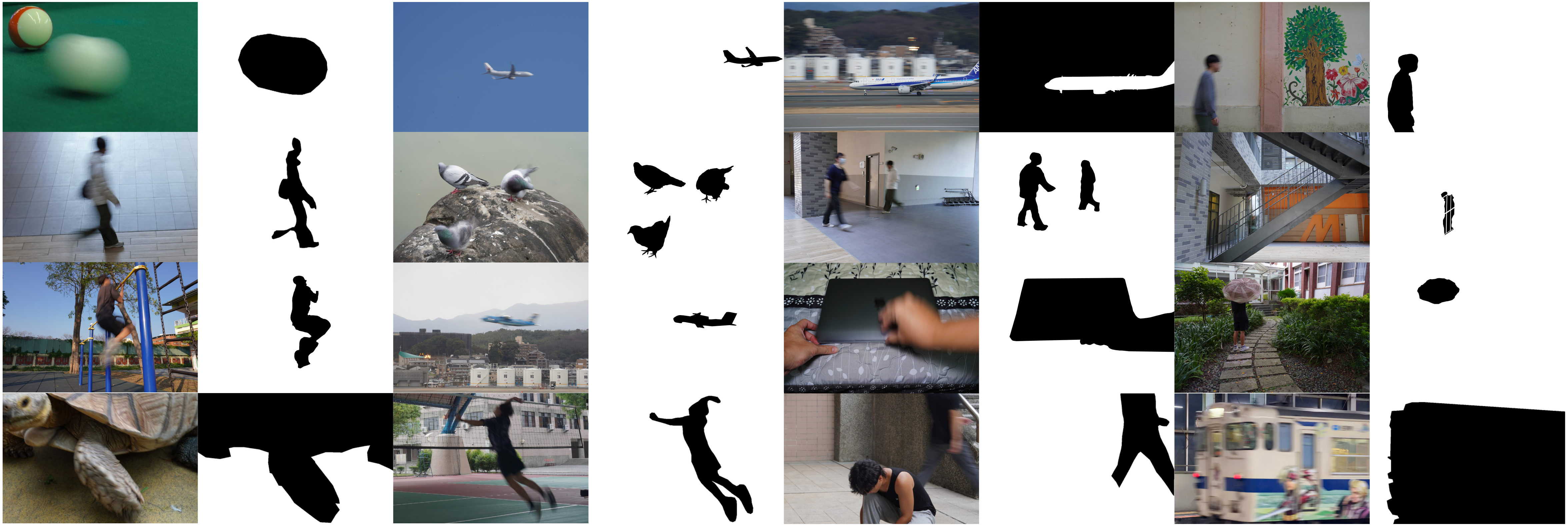}
    \caption{\textbf{Representative samples from BOCCHI (top three rows) and the BOCCHI-protocol subset of the Inference Dataset (bottom row, 164 held-out images for cross-dataset evaluation).} Scenes span pedestrians, vehicles, animals, sports, tabletop manipulation, and rail transit.}
    \label{fig:dataset_samples}
\end{figure}

\textbf{Annotation.} All 633 images were annotated using the LabelMe toolkit~\cite{russell2008labelme} with polygon markups. Each image received three independent candidate annotations, and the most coherent one was selected by majority agreement among annotators. Annotators were instructed to include all visible motion artifacts within the blurred region boundary rather than a conservative interior contour. Representative samples are shown in \cref{fig:dataset_samples}.

\subsection{Dataset Characteristics}

\begin{figure}[!t]
    \centering
    \includegraphics[width=\linewidth]{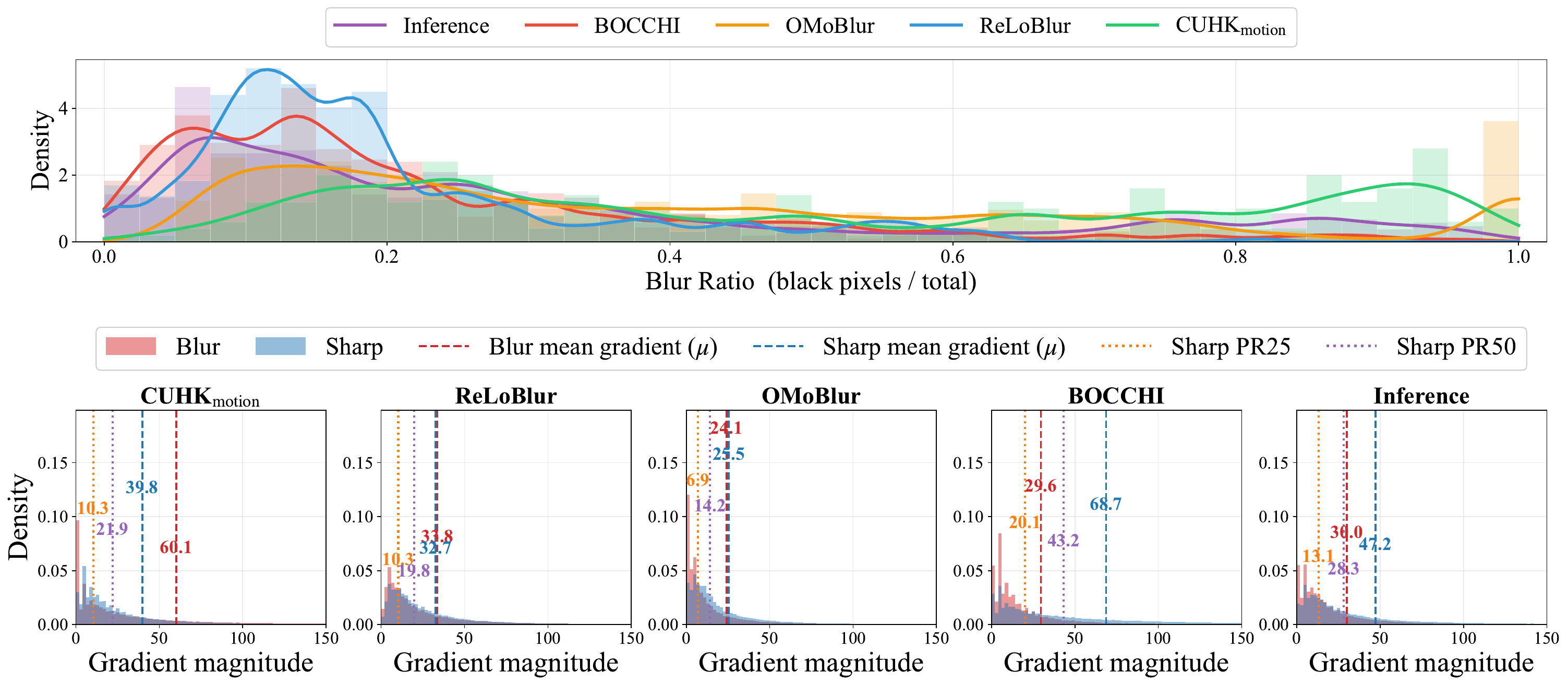}
    \caption{\textbf{Two distinguishing properties of BOCCHI across five datasets.} \emph{(a)} Blur-ratio distribution. \emph{(b)} Per-dataset gradient distribution in blur vs.\ sharp regions, with sharp-region PR25 and PR50 overlays. Quantitative comparisons and the PR25/$\mu_\text{blur}$ ratio are analyzed in \cref{sec:dataset}.}
    \label{fig:bocchi_distinct}
\end{figure}

\cref{tab:dataset_stats} summarizes the statistical characteristics of the five datasets analyzed in this work. We then visualize the two quantities that most distinguish BOCCHI from prior benchmarks (\cref{fig:bocchi_distinct}): the blur-ratio distribution (panel a) and the per-image gradient statistics (panel b).

\begin{table}[t]
\centering
\small
\caption{\textbf{Dataset statistics comparison across five datasets.} BOCCHI and Inference Dataset are established in this work. $Grad_{blur}$/$Grad_{sharp}$ denote per-image mean Sobel gradient magnitude over the corresponding regions; PR25 and PR50 are the 25th and 50th percentiles of sharp-region gradients. Mask convention: black $\to$ blur, white $\to$ sharp.}
\label{tab:dataset_stats}
\vspace{3pt}

\resizebox{\linewidth}{!}{%
\begin{tabular}{l c c c c c c}
\toprule
Dataset & Samples & Blur Ratio & $Grad_{\mathrm{blur}}$ & $Grad_{\mathrm{sharp}}$ & $Sharp_{\mathrm{PR25}}$ & $Sharp_{\mathrm{PR50}}$ \\
\midrule
CUHK$_{\text{motion}}$~\cite{shi2014discriminative} & 200   & 0.51 $\pm$ 0.30 & 60.12 & 39.75 & 10.35 & 21.94 \\
ReLoBlur~\cite{li2023reloblur}                     & 1,200 & 0.20 $\pm$ 0.14 & 33.76 & 32.67 & 10.30 & 19.76 \\
OMoBlur~\cite{yu2026omoblur}                       & 994   & 0.41 $\pm$ 0.28 & 24.09 & 25.47 & 6.87  & 14.18 \\
BOCCHI (Ours)                & 633   & 0.22 $\pm$ 0.19 & 29.62 & \textbf{68.68} & \textbf{20.15} & \textbf{43.22} \\
Inference (Ours)             & 572   & 0.28 $\pm$ 0.25 & 30.00 & 47.17 & 13.14 & 28.29 \\
\bottomrule\end{tabular}
}
\end{table}

\textbf{Blur-ratio coverage.} As shown in panel (a), the training datasets exhibit different blur-ratio distributions: CUHK$_{\text{motion}}$ spans uniformly across the full range with only 200 samples; ReLoBlur ($0.20{\pm}0.14$) and BOCCHI ($0.22{\pm}0.19$) both concentrate at low ratios, representing scenes dominated by small-area blur subjects against larger sharp backgrounds; OMoBlur ($0.41{\pm}0.28$) occupies a moderate range from its burst-averaging synthesis. The Inference Dataset ($0.28{\pm}0.25$) spans the intermediate regime. While ReLoBlur and BOCCHI share similar blur ratios, they differ markedly in their sharp-region gradient statistics (panel b), as discussed next.

\textbf{Sharp-region distribution overlap.} Beyond blur-ratio coverage, BOCCHI exhibits a distinctive sharp-region gradient pattern (panel b). Its sharp region has the highest per-image mean gradient among the five datasets ($\mu_\text{sharp}{=}68.7$), substantially above its blur-region mean ($\mu_\text{blur}{=}29.6$, $\lvert\Delta\rvert{=}39.1$, the largest gap). Yet its bottom-25\% sharp gradient ($\text{PR25}{=}20.2$) falls below the blur mean, giving a PR25/$\mu_\text{blur}$ ratio of $0.68$, the largest among the five datasets (CUHK$_\text{motion}$ $0.17$, ReLoBlur $0.31$, OMoBlur $0.29$, Inference $0.44$). This reflects BOCCHI's bimodal sharp distribution: a high-gradient mode from textured urban backgrounds plus a low-gradient tail from smooth surfaces (sky, plain walls, defocused-yet-sharp regions). Such overlap with the blur distribution makes gradient magnitude alone insufficient to discriminate blur from sharp, motivating our DCT-based design (\cref{sec:method}) that examines frequency content rather than raw spatial gradient.

To support a balanced cross-dataset evaluation in which no single source dominates, we further construct an \textbf{Inference Dataset} of 572 images assembled from four sources of comparable size: (i) 164 newly captured images following the BOCCHI protocol (SONY $\alpha$6400 with E PZ 16-50mm or E 28-200mm, iPhone 17 Pro), (ii) 162 held-out samples from the ReLoBlur~\cite{li2023reloblur} test split, (iii) 96 CUHK$_{\text{motion}}$~\cite{shi2014discriminative} images that are not part of its original training set, and (iv) 150 OMoBlur~\cite{yu2026omoblur} test images for comparison against contemporaneous synthetic local motion blur. All Inference Dataset images are completely excluded from any model training procedures in this paper.

%% file: 04_method.tex
\section{Proposed Method}
\label{sec:method}

\subsection{Overview}

We formulate the task of local motion blur detection as a binary semantic segmentation problem: given an input image $I \in \mathbb{R}^{H \times W \times 3}$, our goal is to predict a pixel-level mask $S \in [0,1]^{H \times W}$.

\begin{figure}[!t]
    \centering
    \includegraphics[width=1\linewidth]{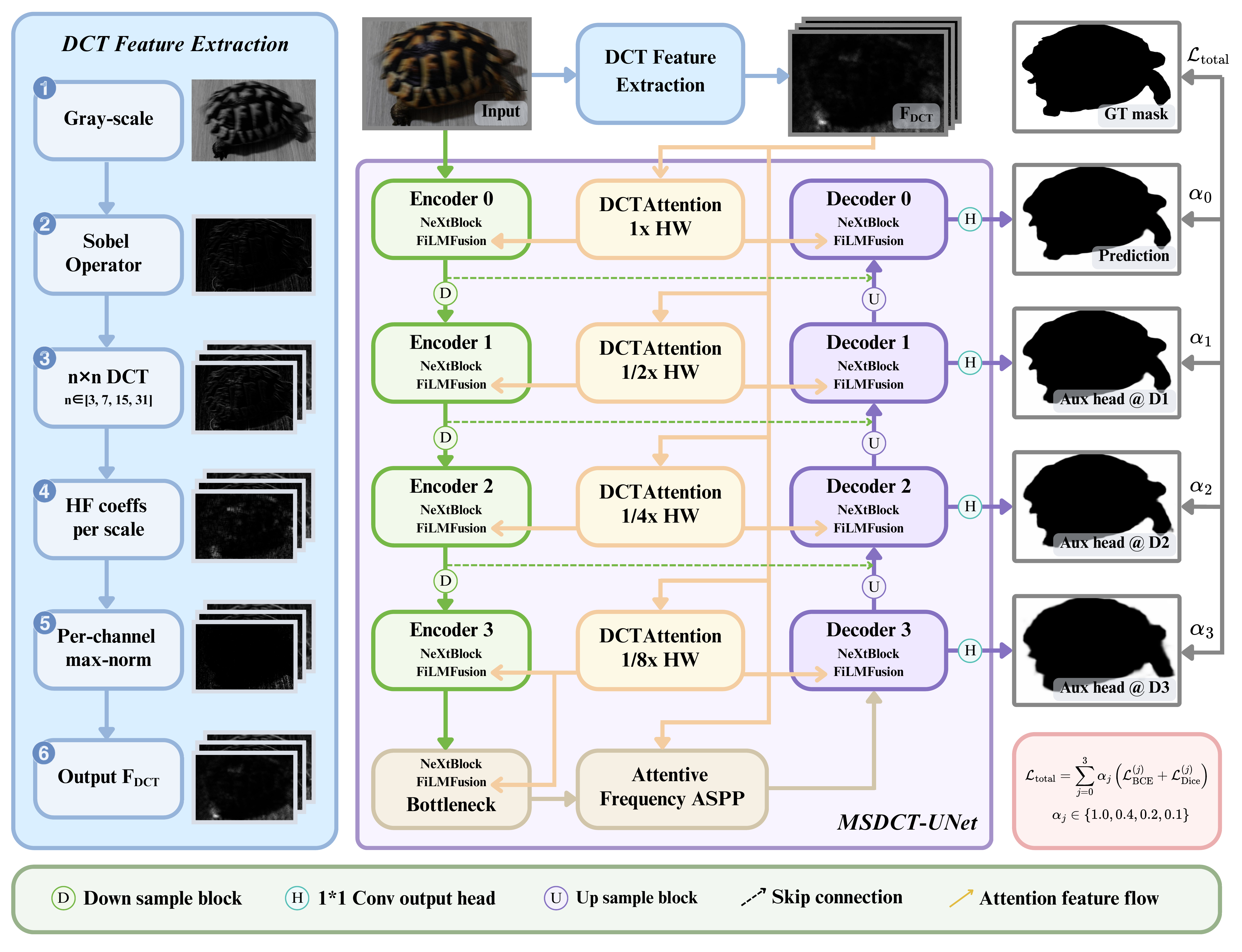}
    \caption{\textbf{Overview of MSDCT-UNet.} \emph{(Left)} DCT feature extraction; \emph{(Middle)} 4-stage encoder-decoder with NeXtBlock+FiLM fusion and an AFASPP bottleneck; \emph{(Right)} deep supervision over main and three auxiliary heads.}
    \label{fig:architecture}
\end{figure}

The harder regime exposed by BOCCHI (\cref{sec:dataset}), where textured foregrounds may be blurred and smooth backgrounds may be sharp, makes purely spatial-domain features insufficient for reliable local blur detection. A model must explicitly disentangle the spectral signature of motion from raw gradient magnitude. Among several candidate frequency decompositions compared in the supplementary material, we adopt the DCT for its spatial locality, real-valued energy-concentrating coefficients, and alignment with the classical High-frequency multi-scale Fusion and Sort Transform (HiFST) prior~\cite{golestaneh2017hifst}. Motivated by this analysis, we propose \textbf{MSDCT-UNet}, which consists of (i) a rule-based DCT feature extraction branch that transforms image gradients into a 57-channel frequency representation (\cref{subsec:dct_ext}), and (ii) a learning-based encoder-decoder network that fuses these frequency priors into every stage through DCT Attention, FiLM modulation, and an Attentive Frequency Atrous Spatial Pyramid Pooling (AFASPP) bottleneck (\cref{subsec:dctnet_arch}). The overall design is illustrated in \cref{fig:architecture}.

\subsection{DCT Feature Extraction}
\label{subsec:dct_ext}

Motion blur causes the progressive attenuation of high-frequency components in images. To effectively capture and exploit this intrinsic property, our preprocessing pipeline is partially based on HiFST~\cite{golestaneh2017hifst}: we adopt its multi-scale local-DCT decomposition and high-frequency selection on gradient magnitudes, but deliberately do not employ its full fusion-and-sort post-processing.

Our procedure first converts the RGB image to grayscale and computes the Sobel gradient magnitude $G$. Following HiFST~\cite{golestaneh2017hifst}, we apply block-wise 2D DCT on $G$ at four different scales $M \in \{3, 7, 15, 31\}$. For a block size $M$, the orthonormal DCT basis matrix $D_M \in \mathbb{R}^{M \times M}$ is defined as
\begin{equation}
    D_M[i,j] = \alpha_i \sqrt{\frac{2}{M}} \cos\!\left( \frac{\pi (2j+1) i}{2M} \right),
    \quad \alpha_i = \begin{cases} 1/\sqrt{2}, & i = 0 \\ 1, & i > 0 \end{cases}
    \label{eq:dct_basis}
\end{equation}
For each pixel $(x,y)$, let $P_M(x,y) \in \mathbb{R}^{M \times M}$ denote the $M \times M$ patch of $G$ centered at $(x,y)$. The 2D DCT coefficient block is
\begin{equation}
    C_M(x,y) = \big| D_M \, P_M(x,y) \, D_M^{\top} \big|.
    \label{eq:dct_2d}
\end{equation}
Together, \cref{eq:dct_basis,eq:dct_2d} produce the per-pixel multi-scale DCT spectrum from the gradient map. At each scale we discard the upper-left low-frequency region of $C_M$ and keep only the high-frequency (HF) positions, yielding $6$, $28$, $120$, and $496$ candidates for $M = 3, 7, 15, 31$ respectively, totaling $650$ HF magnitudes per pixel. From this multi-scale pool we then adaptively select the $57$ smallest magnitudes per pixel via partial sorting, producing a $57$-channel feature map $F_{\text{DCT}} \in \mathbb{R}^{H' \times W' \times 57}$ with $H' = H/4$, $W' = W/4$. The intuition behind selecting the smallest values follows HiFST~\cite{golestaneh2017hifst}: in sharp regions even the weakest HF coefficients retain non-trivial energy, whereas in blurred regions the weakest HF coefficients collapse toward zero, so the per-pixel minima provide a stable indicator of local blur strength. Each channel is then independently max-normalized to $[0, 1]$, and $F_{\text{DCT}}$ is bilinearly resized to match the spatial resolution of each encoder and decoder stage. Further details, including the exact HF position masks and a comparison with fixed-position alternatives, are provided in the supplementary material.

\subsection{MSDCT-UNet Architecture}
\label{subsec:dctnet_arch}

\begin{figure}[t]
    \centering
    \includegraphics[width=\linewidth]{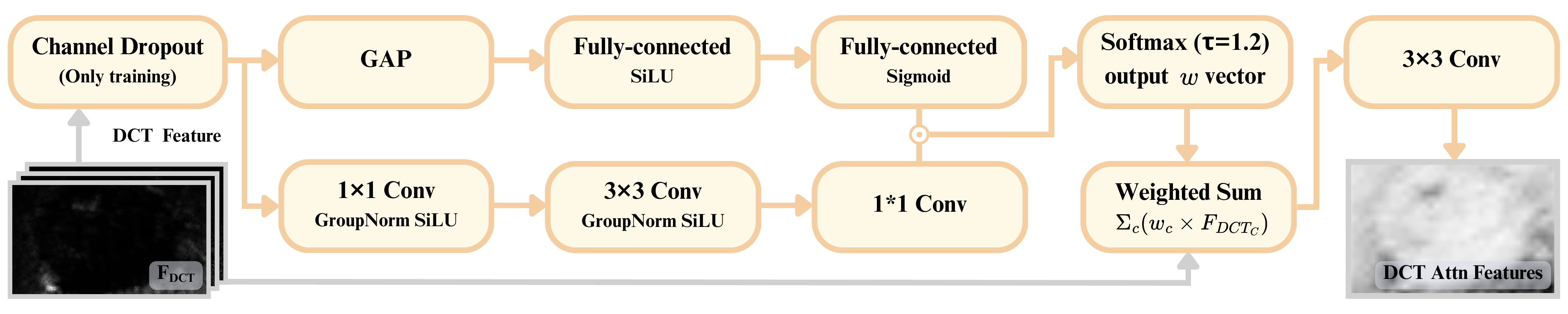}
    \caption{\textbf{DCT Attention (single head).} Each resolution uses $K{=}4$ independent heads with channel-wise dropout ($p{=}0.1$) at training only.}
    \label{fig:dctattn}
\end{figure}

\textbf{DCT Attention.} The 57 DCT channels correspond to coefficients at different scales and frequency positions, but not all channels are equally informative for blur detection. We design a multi-head DCT Attention module that learns position-dependent frequency weighting (\cref{fig:dctattn}). Throughout this section, we use the SiLU activation defined as
\begin{equation}
    \text{SiLU}(x) = x \cdot \sigma(x), \quad \text{where} \quad \sigma(x) = \frac{1}{1+e^{-x}}.
    \label{eq:silu}
\end{equation}
Given DCT features $F_{\text{DCT}} \in \mathbb{R}^{B \times 57 \times H \times W}$, the module operates as follows.

\textbf{Feature refinement.} The features $F_{\text{DCT}}$ are projected to a hidden dimension $d_h{=}48$ through a $1{\times}1$ convolution and then passed through a $3{\times}3$ convolution (each with GroupNorm and SiLU; \cref{eq:silu}) to capture local spatial features, producing $F'$.

\textbf{Multi-head projection.} The feature $F'$ is projected by a single $1{\times}1$ convolution into $K{=}4$ parallel heads, where each head produces 57 channels of logits, denoted as $L \in \mathbb{R}^{B \times K \times 57 \times H \times W}$.

\textbf{SE-gate.} Channel gating values $g \in \mathbb{R}^{57}$ are computed through global average pooling on $F_{\text{DCT}}$ followed by two fully connected layers with sigmoid activation, and broadcast across the $K$ heads to produce gated logits $L^g = L \odot g$.

\textbf{Temperature-scaled softmax and weighted sum.} For each head, a softmax with temperature $\tau{=}1.2$ is applied along the 57 channel dimension of $L^g$ to produce smoother attention weights, which are then used to linearly combine the original DCT channels into a single activation map per head.

\textbf{Spatial smoothing.} Finally, the attention maps $\{a_k\}$ are passed through a depthwise $3{\times}3$ convolution to obtain the final attention map $A \in \mathbb{R}^{B \times K \times H \times W}$.

During training, channel-wise dropout ($p{=}0.10$) is applied to the DCT input, randomly zeroing out some frequency channels to enhance robustness and improve training stability.

\begin{figure}[t]
    \centering
    \includegraphics[width=1.0\linewidth]{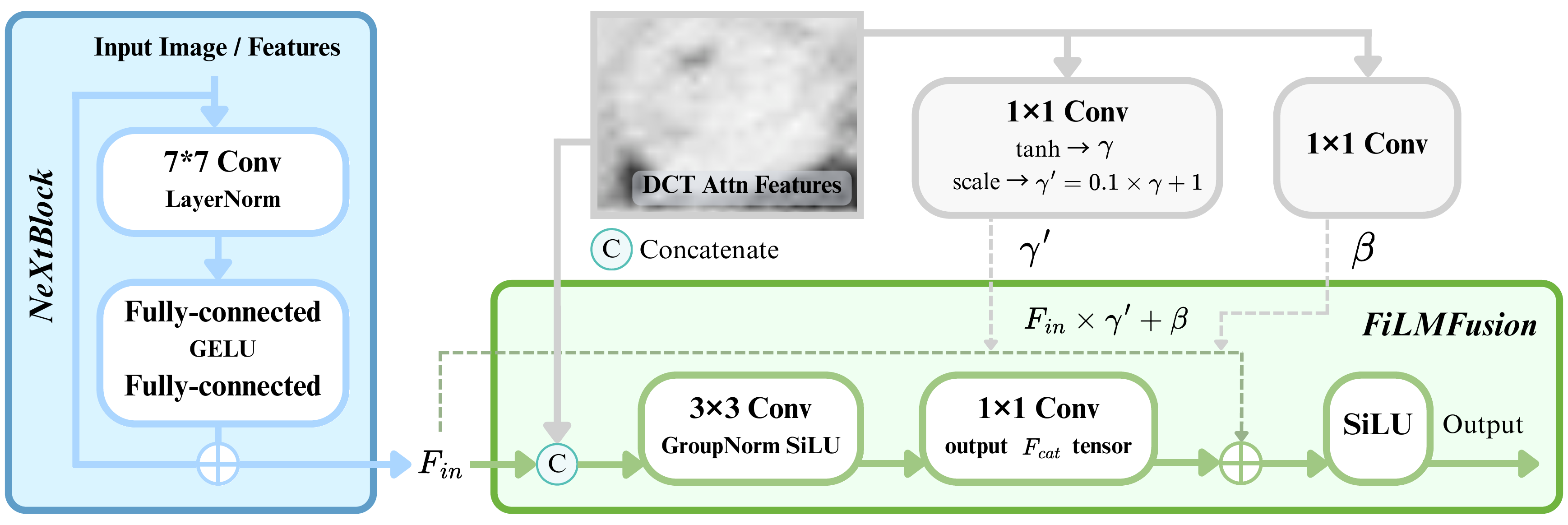}
    \caption{\textbf{NeXtBlock + FiLM Fusion.} The modulation path (dashed) bypasses the concatenation; $F_{\text{in}}$ and the DCT attention map $A$ are combined as in \cref{eq:film}.}
    \label{fig:nextfilm}
\end{figure}

\textbf{FiLM Fusion.} Frequency features encode \emph{where high-frequency components are attenuated} (i.e., blur evidence), while spatial features encode \emph{where intensity changes occur} (i.e., content edges). These two signals are inherently complementary but not directly additive: naively summing them allows strong spatial edges to easily dominate the weaker frequency cues. We therefore adopt FiLM (Feature-wise Linear Modulation)~\cite{perez2018film} for fusion, which adaptively modulates spatial features through attention-conditioned affine transforms. Specifically, we process the spatial features $F_{\text{in}} \in \mathbb{R}^{C \times H \times W}$ and the attention map $A \in \mathbb{R}^{K \times H \times W}$ through two parallel branches.

\textit{Concatenation path.} $F_{\text{in}}$ and $A$ are concatenated along the channel dimension and fused by a $3{\times}3$ convolution (with GroupNorm and SiLU) followed by a $1{\times}1$ convolution, producing $F_{\text{cat}}$.

\textit{FiLM modulation path.} A pair of scale and shift maps $(\gamma, \beta)$ are predicted from $A$ via $1{\times}1$ convolutions, and combined with the spatial features as
\begin{equation}
    F_{\text{out}} = \text{SiLU}\!\left( F_{\text{cat}} + F_{\text{in}} \cdot \big(1 + 0.1 \cdot \tanh(\gamma)\big) + \beta \right),
    \label{eq:film}
\end{equation}
where $\tanh(\gamma)$ bounds the scale range and the factor $0.1$ keeps the modulation close to an identity mapping at the beginning of training.

\textbf{Encoder-Decoder Backbone.} The backbone follows a four-stage symmetric U-Net design with channel widths 64, 128, 256, and 512. At each encoder stage we apply a NeXtBlock~\cite{liu2022convnext} followed by FiLM fusion, with stride-2 $3{\times}3$ convolutions for downsampling between stages. The decoder is symmetric: each block applies bilinear upsampling, concatenates the skip connection from the corresponding encoder stage, and passes the result through another NeXtBlock + FiLM fusion. Each of the four resolutions owns an \emph{independent} DCTAttention instance (four in total, with non-shared weights): the 57-channel $F_{\text{DCT}}$ map is bilinearly resized to that stage's spatial resolution and fed into the stage's DCTAttention to produce the attention map $A$, consumed by both the encoder and the corresponding decoder block at the same resolution.

\begin{figure}[t]
    \centering
    \includegraphics[width=\linewidth]{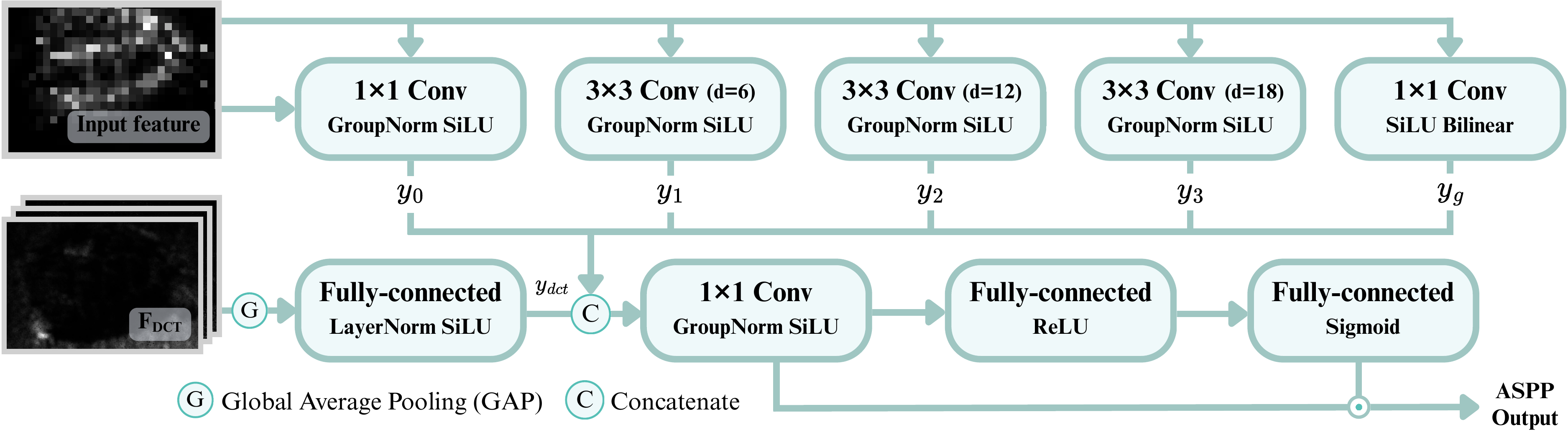}
    \caption{\textbf{Attentive Frequency ASPP.} Five spatial branches and a DCT projection branch concatenated and SE-recalibrated.}
    \label{fig:afaspp}
\end{figure}

\textbf{Attentive Frequency ASPP.} At the bottleneck between encoder and decoder, we introduce AFASPP (\cref{fig:afaspp}), which fuses multi-scale spatial context with global frequency evidence. AFASPP contains five spatial branches (a $1{\times}1$ convolution, three dilated $3{\times}3$ convolutions with rates $\{6, 12, 18\}$, and global average pooling) plus one DCT projection branch that linearly projects the globally pooled $F_{\text{DCT}}$ vector into the spatial embedding. The six branches are then concatenated along the channel dimension and projected by a $1{\times}1$ convolution, followed by an SE Block~\cite{hu2018senet} that adaptively recalibrates the channels based on global feature statistics.

\textbf{Deep Supervision.} In addition to the main output, a $1{\times}1$ convolution auxiliary head is placed at each of the decoder's D1/D2/D3 stages, with supervision weights $0.4$, $0.2$, and $0.1$ respectively.

\subsection{Loss Function}

The loss function adopts a combination of Binary Cross-Entropy (BCE) and Dice Loss. Given predicted logits $z \in \mathbb{R}^{B \times 1 \times H \times W}$, the per-pixel probability is $p = \sigma(z)$, with $\sigma(\cdot)$ defined in \cref{eq:silu}. With ground-truth mask $y \in \{0,1\}^{B \times 1 \times H \times W}$, the BCE loss with positive-class weighting is:
\begin{equation}
    \mathcal{L}_{\text{BCE}} = -\frac{1}{N} \sum_{i} \big[ w_p \cdot y_i \log p_i + (1 - y_i) \log(1 - p_i) \big]
    \label{eq:bce}
\end{equation}
where $N$ is the total number of pixels in the image and $w_p$ is the positive-class weight automatically estimated from the foreground/background ratio of each training batch (clipped to a maximum of 50.0 for stability). The Dice loss is computed per channel and averaged:
\begin{equation}
    \mathcal{L}_{\text{Dice}} = 1 - \frac{1}{C} \sum_{c=1}^{C} \frac{2 \sum_{b,h,w} p_{b,c,h,w} \, y_{b,c,h,w} + \epsilon}{\sum_{b,h,w} p_{b,c,h,w} + \sum_{b,h,w} y_{b,c,h,w} + \epsilon}
    \label{eq:dice}
\end{equation}
where $\epsilon = 10^{-6}$ avoids division by zero and the summation reduces over the batch and spatial dimensions.

With deep supervision, the network produces a main output $z^{(0)}$ and three auxiliary outputs $z^{(1)}, z^{(2)}, z^{(3)}$ from intermediate decoder stages D1, D2, D3. The auxiliary outputs are upsampled to the target resolution before the loss computation. The total loss is:
\begin{equation}
    \mathcal{L}_{\text{total}} = \sum_{j=0}^{3} \alpha_j \left( \mathcal{L}_{\text{BCE}}^{(j)} + \mathcal{L}_{\text{Dice}}^{(j)} \right)
    \label{eq:total_loss}
\end{equation}
where $\alpha_j \in \{1.0, 0.4, 0.2, 0.1\}$ are the deep supervision weights assigned for the main and three auxiliary outputs, respectively. Following common deep supervision practice~\cite{zhao2017pspnet}, we deliberately assign smaller weights to lower-resolution auxiliary outputs so that intermediate stages provide gradient guidance without overwhelming the main loss.

%% file: 05_experiment.tex
\section{Experiments}
\label{sec:experiments}

\subsection{Implementation Details}

We train at $270 \times 180$ resolution ($1/4$ of $1080 \times 720$) with BCE + Dice loss and deep supervision (\cref{eq:bce,eq:dice,eq:total_loss}). Splits are BOCCHI (570/63), CUHK$_{\text{motion}}$~\cite{shi2014discriminative} (180/20), ReLoBlur~\cite{li2023reloblur} (1080/120), and OMoBlur~\cite{yu2026omoblur} (994/100); ReLoBlur and OMoBlur are randomly subsampled to comparable scales as BOCCHI. Cross-dataset transfer is evaluated on the Inference Dataset (572 images, with 28.7/28.3/26.2/16.8\% drawn from BOCCHI/ReLoBlur/OMoBlur/CUHK$_{\text{motion}}$). Metrics: mean IoU (mIoU), Dice (F1), Boundary F1 (BdF1, $\tau{=}2$ px). We compare MSDCT-UNet against 12 baselines spanning lightweight real-time segmentation~\cite{yu2021bisenetv2,fan2021stdc,hong2022ddrnet,msdsegnet,li2025esmdlunet}, multi-scale architectures~\cite{msdunet}, and some recent methods~\cite{chen2022nafnet,kdsnet,stringer2021cellpose,zhao2025bevanet}, all under identical training settings. Full hyperparameters and hardware details are in the supplementary.

\subsection{Comparison with State-of-the-Art}

\begin{figure}[t]
    \centering
    \includegraphics[width=\linewidth]{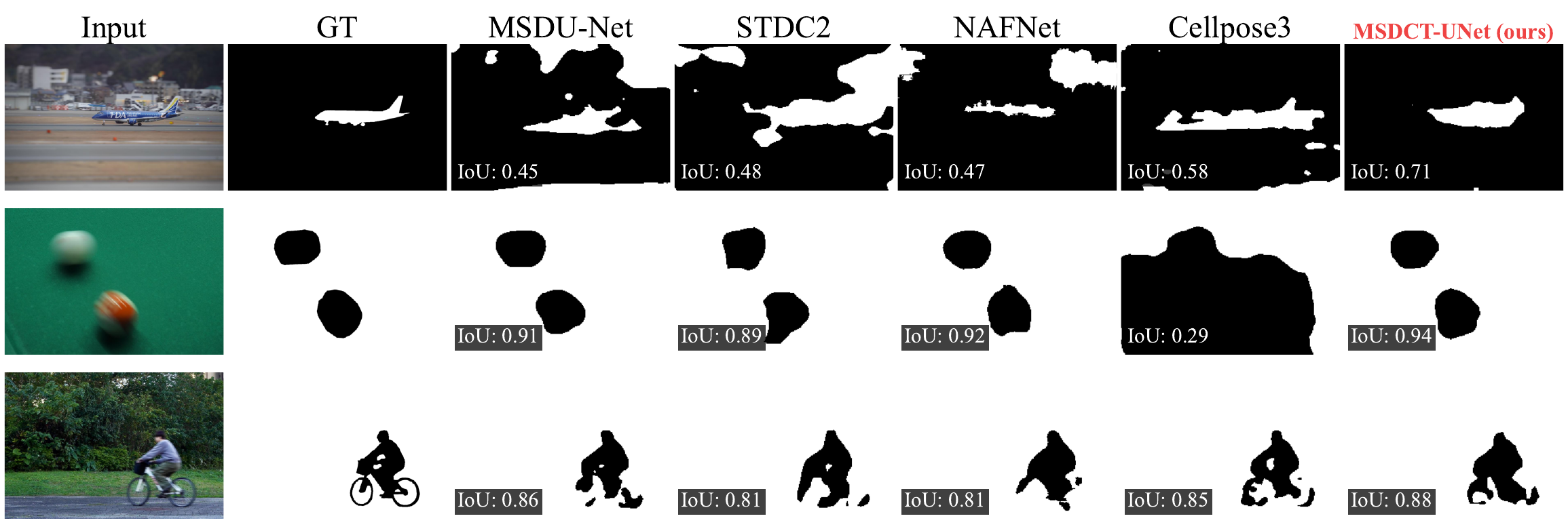}
    \caption{\textbf{Qualitative comparison on the BOCCHI validation set.} Three representative cases. Predictions are binary masks (black = predicted blur). Full 12-baseline grids and failure cases are in the supplementary.}
    \label{fig:qualitative}
\end{figure}

\cref{tab:indomain_all} reports in-domain results across 13 models on four datasets. On BOCCHI, MSDCT-UNet ranks first in both metrics, with the largest margin on BdF1 ($+4.0$ pp over Cellpose3~\cite{stringer2021cellpose}), confirming that frequency cues particularly aid boundary localization. On ReLoBlur it ranks first in both metrics; on CUHK$_{\text{motion}}$, tied second in mIoU (with MSDU-Net) and second in BdF1; on OMoBlur~\cite{yu2026omoblur}, second in mIoU and third in BdF1, trailing Cellpose3 on this synthetic source. A paired per-image comparison on the BOCCHI validation split further shows that the improvement is not concentrated on a few outliers: MSDCT-UNet outperforms Cellpose3 on $38$ of $63$ images ($60.3\%$), with a median per-image mIoU improvement of $+0.7$ pp. Although we do not claim statistical significance without repeated training, this paired analysis indicates that MSDCT-UNet's advantage is broadly distributed rather than driven by isolated cases. Qualitative comparisons on representative cases are shown in \cref{fig:qualitative}.

\input{tables/indomain_all}

\input{tables/cross_dataset}

\subsection{BOCCHI as a Discriminative Benchmark}
\label{sec:discriminative}

A frequency-agnostic spatial backbone (DDRNet-23~\cite{hong2022ddrnet}) trails MSDCT-UNet by $2.0$ pp on ReLoBlur but by $14.9$ pp on BOCCHI, a $7.5{\times}$ increase. Spatial-only backbones are more strongly penalized on BOCCHI's gradient-overlap regime (sharp PR25 below blur mean), making BOCCHI a more stringent test of gradient-shortcut dependence.

\subsection{Data Efficiency Analysis}
\label{sec:data_efficiency}

\cref{tab:cross_dataset} reports cross-dataset transfer (no fine-tuning). The Average row aggregates 13 models (12 for OMoBlur; NAFNet omitted) and shows BOCCHI-trained models outperform every other source on all three metrics: BOCCHI $0.563$ vs.\ OMoBlur $0.541$ ($+2.2$/$+7.9$/$+15.8$ pp mIoU/Dice/Recall), ReLoBlur $0.425$ ($+13.8$/$+16.8$/$+23.9$ pp), CUHK$_{\text{motion}}$ $0.214$ ($+34.9$ pp mIoU). This is achieved with only $633$ training images vs.\ ReLoBlur ($1{,}200$) and OMoBlur ($994$), and the advantage is architecture-agnostic: $13$/$13$ models follow BOCCHI $\geq$ ReLoBlur on mIoU. The in-domain $\to$ cross-dataset gap ($0.850 \to 0.563$) is discussed in the supplementary.

\input{tables/ablation_combined}

\subsection{Ablation Study}

\cref{tab:ablation_combined} reports two ablations on BOCCHI, both evaluated in-domain. \emph{Architecture:} the Freq Branch is the key component (in-domain mIoU drops $4.6$ pp on removal); adding the Freq Attention on top further raises in-domain mIoU by $0.5$ pp and BdF1 by $2.4$ pp. \emph{Loss:} BCE + Dice achieves the best in-domain mIoU $0.850$ and BdF1 $0.683$; Dice alone performs the worst, indicating BCE is necessary for stable training. Extended ablations (architecture $\times$ loss interaction, frequency-transform comparison, additional analyses) are provided in the supplementary.

%% file: tables/indomain_all.tex

\begin{table}[t]
\centering
\small
\fontsize{8pt}{9.5pt}\selectfont
\caption{\textbf{In-domain results across four datasets.} All models are trained and evaluated using a 90/10 split on each respective dataset. Per-cell metrics: mIoU / BdF1; Dice omitted for space. \textsuperscript{*}NAFNet $\times$ OMoBlur is omitted because OMoBlur's $384{\times}1920$ aspect ratio causes width-dimension collapse at NAFNet's deepest downsampling stage. Results are highlighted as \colorbox{lightred}{1st}, \colorbox{lightorange}{2nd}, and \colorbox{lightyellow}{3rd} for each column.}
\label{tab:indomain_all}
\vspace{3pt}
\resizebox{\textwidth}{!}{%
\begin{tabular}{l | c c | c c | c c | c c}
\toprule
\multicolumn{1}{c|}{}
& \multicolumn{2}{c|}{BOCCHI} & \multicolumn{2}{c|}{CUHK$_{\text{motion}}$} & \multicolumn{2}{c|}{ReLoBlur} & \multicolumn{2}{c}{OMoBlur} \\
\cmidrule(lr){2-3} \cmidrule(lr){4-5} \cmidrule(lr){6-7} \cmidrule(lr){8-9}
Model & mIoU$\uparrow$ & BdF1$\uparrow$ & mIoU$\uparrow$ & BdF1$\uparrow$ & mIoU$\uparrow$ & BdF1$\uparrow$ & mIoU$\uparrow$ & BdF1$\uparrow$ \\
\midrule
MSDU-Net~\cite{msdunet}                  & 0.796 & \cellcolor{lightyellow}0.561 & \cellcolor{lightorange}0.640 & \cellcolor{lightyellow}0.330 & \cellcolor{lightyellow}0.923 & \cellcolor{lightyellow}0.820 & \cellcolor{lightyellow}0.772 & \cellcolor{lightorange}0.667 \\
BiSeNetV2~\cite{yu2021bisenetv2}         & 0.760 & 0.470 & 0.525 & 0.166 & 0.916 & 0.800 & 0.749 & 0.613 \\
STDC1~\cite{fan2021stdc}                 & 0.778 & 0.457 & 0.550 & 0.260 & 0.918 & 0.795 & 0.748 & 0.631 \\
STDC2~\cite{fan2021stdc}                 & \cellcolor{lightyellow}0.799 & 0.495 & 0.558 & 0.197 & 0.918 & 0.800 & 0.755 & 0.627 \\
NAFNet~\cite{chen2022nafnet}             & 0.760 & 0.452 & 0.523 & 0.213 & 0.914 & 0.821 & N/A\textsuperscript{*} & N/A\textsuperscript{*} \\
DDRNet-23~\cite{hong2022ddrnet}          & 0.701 & 0.351 & 0.520 & 0.172 & 0.910 & 0.766 & 0.703 & 0.583 \\
Cellpose3~\cite{stringer2021cellpose}    & \cellcolor{lightorange}0.838 & \cellcolor{lightorange}0.643 & \cellcolor{lightred}0.665 & \cellcolor{lightred}0.403 & \cellcolor{lightorange}0.927 & \cellcolor{lightorange}0.850 & \cellcolor{lightred}0.791 & \cellcolor{lightred}0.675 \\
KDSNet-R50~\cite{kdsnet}                 & 0.783 & 0.469 & 0.549 & 0.219 & 0.918 & 0.803 & 0.748 & 0.627 \\
KDSNet-R101~\cite{kdsnet}                & 0.773 & 0.446 & 0.544 & 0.243 & 0.921 & 0.812 & 0.744 & 0.630 \\
MSDSeg~\cite{msdsegnet}                  & 0.714 & 0.360 & 0.517 & 0.202 & 0.909 & 0.766 & 0.721 & 0.607 \\
BEVANet~\cite{zhao2025bevanet}           & 0.778 & 0.518 & 0.548 & 0.217 & 0.919 & 0.798 & 0.721 & 0.622 \\
ESMDL-UNet~\cite{li2025esmdlunet}        & 0.677 & 0.270 & 0.432 & 0.139 & 0.788 & 0.364 & 0.627 & 0.552 \\
\midrule
MSDCT-UNet (Ours)                        & \cellcolor{lightred}0.850 & \cellcolor{lightred}0.683 & \cellcolor{lightorange}0.640 & \cellcolor{lightorange}0.369 & \cellcolor{lightred}0.930 & \cellcolor{lightred}0.851 & \cellcolor{lightorange}0.775 & \cellcolor{lightyellow}0.664 \\
\bottomrule
\end{tabular}%
}

\end{table}

%% file: tables/cross_dataset.tex

\begin{table}[t]
\centering
\small
\fontsize{8pt}{9.5pt}\selectfont
\caption{\textbf{Cross-dataset generalization.} Per-cell metrics: mIoU / Dice / Recall on the Inference Dataset (572 images, no fine-tuning). Parentheses are source dataset sizes. The \textit{Average} row highlights BOCCHI as the strongest training source on all three reported metrics. \textsuperscript{*}NAFNet $\times$ OMoBlur is omitted because OMoBlur's $384{\times}1920$ aspect ratio causes width-dimension collapse at NAFNet's deepest downsampling stage (width $= 1$); OMoBlur Average aggregates 12 of 13 models. Per-cell Precision, Boundary-F1, and Boundary-IoU are in the supplementary material.}
\label{tab:cross_dataset}
\vspace{3pt}
\resizebox{\textwidth}{!}{
\begin{tabular}{l c c c c}
\toprule
\multirow{2}{*}{Model} & \multicolumn{4}{c}{Training source (mIoU / Dice / Recall)} \\
\cmidrule(lr){2-5}
 & BOCCHI (633) & OMoBlur (994) & ReLoBlur (1{,}200) & CUHK$_{\text{motion}}$ (200) \\
\midrule
MSDU-Net~\cite{msdunet}                   & 0.609 / 0.830 / 0.876 & 0.616 / 0.805 / 0.810 & 0.556 / 0.758 / 0.736 & 0.142 / 0.307 / 0.268 \\
BiSeNetV2~\cite{yu2021bisenetv2}          & 0.539 / 0.791 / 0.843 & 0.526 / 0.704 / 0.666 & 0.470 / 0.695 / 0.651 & 0.248 / 0.445 / 0.455 \\
STDC1~\cite{fan2021stdc}                  & 0.541 / 0.789 / 0.840 & 0.496 / 0.657 / 0.597 & 0.283 / 0.518 / 0.538 & 0.197 / 0.310 / 0.286 \\
STDC2~\cite{fan2021stdc}                  & 0.577 / 0.823 / 0.882 & 0.530 / 0.705 / 0.655 & 0.269 / 0.443 / 0.437 & 0.182 / 0.288 / 0.277 \\
NAFNet~\cite{chen2022nafnet}              & 0.568 / 0.818 / 0.882 & N/A\textsuperscript{*}        & 0.478 / 0.730 / 0.718 & 0.285 / 0.529 / 0.533 \\
DDRNet-23~\cite{hong2022ddrnet}           & 0.517 / 0.800 / 0.877 & 0.489 / 0.673 / 0.626 & 0.462 / 0.704 / 0.665 & 0.292 / 0.548 / 0.490 \\
Cellpose3~\cite{stringer2021cellpose}     & 0.662 / 0.858 / 0.910 & 0.628 / 0.825 / 0.848 & 0.555 / 0.758 / 0.744 & 0.160 / 0.324 / 0.306 \\
KDSNet-R50~\cite{kdsnet}                  & 0.557 / 0.809 / 0.871 & 0.528 / 0.717 / 0.684 & 0.275 / 0.465 / 0.465 & 0.225 / 0.424 / 0.369 \\
KDSNet-R101~\cite{kdsnet}                 & 0.557 / 0.794 / 0.835 & 0.525 / 0.718 / 0.696 & 0.274 / 0.433 / 0.423 & 0.230 / 0.432 / 0.372 \\
MSDSeg~\cite{msdsegnet}                   & 0.518 / 0.816 / 0.901 & 0.547 / 0.788 / 0.828 & 0.463 / 0.726 / 0.711 & 0.141 / 0.128 / 0.121 \\
BEVANet~\cite{zhao2025bevanet}            & 0.552 / 0.804 / 0.861 & 0.535 / 0.746 / 0.731 & 0.484 / 0.718 / 0.692 & 0.271 / 0.493 / 0.434 \\
ESMDL-UNet~\cite{li2025esmdlunet}         & 0.518 / 0.806 / 0.873 & 0.458 / 0.698 / 0.684 & 0.391 / 0.631 / 0.593 & 0.231 / 0.397 / 0.352 \\
\midrule
MSDCT-UNet (Ours)                         & 0.610 / 0.848 / 0.925 & 0.610 / 0.783 / 0.781 & 0.563 / 0.815 / 0.897 & 0.183 / 0.290 / 0.226 \\
\midrule
\textit{Average} & \cellcolor{lightred}\textit{0.563 / 0.814 / 0.875} & \cellcolor{lightorange}\textit{0.541 / 0.735 / 0.717}\textsuperscript{*} & \cellcolor{lightyellow}\textit{0.425 / 0.646 / 0.636} & \textit{0.214 / 0.378 / 0.345} \\
\bottomrule
\end{tabular}
}
\vspace{-6pt}
\end{table}

%% file: tables/ablation_combined.tex
\begin{table}[!tb]
\centering
\footnotesize
\caption{\textbf{Ablations on BOCCHI (in-domain).} \emph{Left:} architecture (default loss = BCE+Dice). \emph{Right:} loss (on full MSDCT-UNet).}
\label{tab:ablation_combined}
\vspace{3pt}
\begin{minipage}[t]{0.49\linewidth}
\centering
\resizebox{\linewidth}{!}{%
\begin{tabular}{l | c c c}
\toprule
\multicolumn{4}{l}{\textit{Architecture (BCE+Dice)}} \\
\midrule
Variant & mIoU$\uparrow$ & Dice$\uparrow$ & BdF1$\uparrow$ \\
\midrule
Full (MSDCT-UNet) & \textbf{0.850} & 0.948 & \textbf{0.683} \\
w/o Freq Attn     & 0.845 & \textbf{0.949} & 0.659 \\
w/o Freq Branch   & 0.804 & 0.919 & 0.571 \\
\bottomrule
\end{tabular}
}
\end{minipage}\hfill
\begin{minipage}[t]{0.49\linewidth}
\centering
\resizebox{\linewidth}{!}{%
\begin{tabular}{l | c c c}
\toprule
\multicolumn{4}{l}{\textit{Loss (on full MSDCT-UNet)}} \\
\midrule
Variant & mIoU$\uparrow$ & Dice$\uparrow$ & BdF1$\uparrow$ \\
\midrule
BCE + Dice & \textbf{0.850} & \textbf{0.948} & \textbf{0.683} \\
BCE        & 0.840 & 0.944 & 0.670 \\
Dice       & 0.821 & 0.938 & 0.624 \\
\bottomrule
\end{tabular}
}
\end{minipage}
\vspace{-6pt}
\end{table}

%% file: 10_conclusion.tex
\section{Conclusion}
\label{sec:conclusion}

Existing local motion-blur benchmarks can permit gradient-based shortcuts when blurred and sharp regions are easily separable. \textbf{BOCCHI} has addressed this by introducing real-captured images whose sharp-region gradients overlap strongly with blur-region gradients, making detection less reliant on simple gradient cues. Models trained on BOCCHI generalize better to other datasets despite fewer training images, and our \textbf{MSDCT-UNet}, a frequency-aware encoder-decoder with multi-scale DCT priors, has achieved the best in-domain performance on BOCCHI. We discuss continual and multi-source learning as future directions in the supplementary.

%% file: _supp_body.tex
\section{Implementation Details (Extended)}
\label{sec:supp_impl}

\textbf{Software environment.} All experiments are conducted in PyTorch 2.0 with CUDA 11.8 on a single NVIDIA RTX 5070 Ti GPU.

\textbf{Optimization hyperparameters.} We train for 60 epochs with batch size 4, using AdamW (learning rate $10^{-4}$, weight decay $10^{-4}$) and cosine annealing for the learning rate. Gradients are clipped to L2 norm $1.0$. The total loss is BCE + Dice with deep supervision weights $\alpha_j \in \{1.0, 0.4, 0.2, 0.1\}$ for the main and three auxiliary outputs (see main paper Sec.~4.4).

\textbf{Train/validation splits.} BOCCHI uses a 90/10 split (570 training, 63 validation), CUHK$_{\text{motion}}$ uses 180/20, ReLoBlur uses 1{,}080/120, and OMoBlur uses 994 training images (random-subsampled, see main paper Sec.~5.1). The Inference Dataset (572 images) is held out entirely from training and used only for cross-dataset evaluation.

\textbf{Mixed-precision training.} We adopt automatic mixed-precision (AMP) training with PyTorch's built-in \texttt{GradScaler}, which keeps critical operations (e.g., the loss computation) in FP32 for numerical stability while running the bulk of the forward and backward pass in FP16.

\textbf{Random seed.} For reproducibility, we fix the random seed to $531$ across Python \texttt{random}, NumPy, and PyTorch. The 90\%/10\% train/validation split is also produced with this seed.

\textbf{Data augmentation.} During training we apply synchronized geometric transforms to the RGB image and its annotation mask: random horizontal/vertical flips, random rotations, random scaling, and random crops. Color jitter (brightness, contrast, saturation) is applied only to the RGB image and does not affect the pre-computed DCT features.

\section{DCT Feature Extraction Details}
\label{sec:supp_hifst}

Section~4.2 of the main paper summarizes the HiFST-inspired DCT feature extraction. Here we provide the full specification for reproducibility.

\textbf{High-frequency position masks.} For each scale $M \in \{3, 7, 15, 31\}$, the $M \times M$ DCT coefficient matrix $C_M$ is partitioned into a low-frequency region (the upper-left triangular region around the DC entry, including the DC itself) and a high-frequency (HF) region (all remaining positions). Concretely, position $(i, j)$ is labeled low-frequency if $i + j < \lceil M / 2 \rceil$ or if $(i, j)$ lies on the symmetric counterpart of this triangle along the matrix anti-diagonal; the rest are HF. This partition yields $|\text{HF}_M| = 6, 28, 120, 496$ candidate positions for $M = 3, 7, 15, 31$, totaling $650$ HF magnitudes per pixel after concatenation across the four scales.

\textbf{Per-pixel adaptive selection.} Let $\mathbf{f}(x, y) \in \mathbb{R}^{650}$ denote the concatenated HF magnitude vector at pixel $(x, y)$. We compute
\begin{equation}
    F_{\text{DCT}}(x, y) = \mathbf{f}(x, y)\big[\,\text{argpartition}(\mathbf{f}(x, y), 57)[:57]\,\big],
    \label{eq:supp_argpart}
\end{equation}
that is, the 57 smallest entries of $\mathbf{f}(x, y)$ (returned in arbitrary order by partial sorting). This is the central design choice of HiFST: rather than fixing 57 frequency positions globally, each pixel adaptively selects the 57 weakest HF responses from its own multi-scale candidate pool. As a result, the channel index $k \in \{0, \dots, 56\}$ does \emph{not} correspond to a fixed $(M, i, j)$ triple: at one pixel channel $5$ may originate from the $7 \times 7$ scale, while at another pixel it may come from the $31 \times 31$ scale.

\textbf{Why select the smallest magnitudes?} The selection criterion follows the HiFST blur prior: in sharp regions, motion blur has not attenuated high-frequency content, and even the weakest of the $650$ HF coefficients retains non-trivial energy; in motion-blurred regions, high-frequency content is systematically suppressed along the blur direction, so the weakest HF coefficients collapse toward zero. The per-pixel minima thus form a robust low-energy envelope of the HF spectrum, monotonically decreasing with local blur strength, which provides a more discriminative signal than the maximum (which is dominated by occasional residual edges) or the mean (which is biased by texture density).

\textbf{Channel count.} The output dimension $57 = 1 + (3 + 7 + 15 + 31)$ is inherited from the original HiFST formulation~\cite{golestaneh2017hifst} as a multi-scale design constant. Despite the additive form, the leading $+1$ does not correspond to a separate DC channel in our implementation: all 57 channels are HF magnitudes selected as in \cref{eq:supp_argpart}.

\textbf{Per-channel max-normalization.} After argpartition, the resulting feature map $L \in \mathbb{R}^{H' W' \times 57}$ (with $H' = H/4$, $W' = W/4$ from a downsampling stride of $4$) is normalized independently per channel: $L[:, k] \leftarrow L[:, k] / \max(L[:, k])$ for each $k$, ensuring all channels share a comparable $[0, 1]$ dynamic range before being passed to the DCT Attention module. The resulting $F_{\text{DCT}} \in \mathbb{R}^{H' \times W' \times 57}$ is computed once per image at preprocessing time and bilinearly resized inside the network to match each encoder/decoder stage.

\section{Frequency Transform Ablation}
\label{sec:supp_freq}

\textbf{Why DCT?} The main paper adopts DCT as its frequency-domain prior following the design of HiFST~\cite{golestaneh2017hifst}. Here we ablate this choice against ten alternative frequency transforms spanning three families: FFT with varying number of radial bands (4/8/16), orthogonal discrete wavelets at multiple decomposition levels (DWT-db4-L1/L2/L4 and DWT-Haar-L4), and stationary (shift-invariant) wavelets (SWT-db4-L1/L2/L3). For each alternative we keep the remainder of the MSDCT-UNet architecture fixed and replace only the feature-extraction branch; the number of frequency channels therefore varies with the transform (from 3 to 57). All variants are trained with the final BCE+Dice objective on BOCCHI and evaluated on the BOCCHI validation set.

\begin{table}[t]
\centering
\small
\caption{\textbf{Frequency transform ablation on BOCCHI.} Each row reports an alternative transform substituted for DCT in the MSDCT-UNet frequency branch. \textbf{Bold} marks the best value per column.}
\label{tab:supp_freq}
\begin{tabular}{l c c c c}
\toprule
Transform & Ch. & mIoU$\uparrow$ & Dice$\uparrow$ & BdF1$\uparrow$ \\
\midrule
DCT (Ours)        & 57 & \textbf{0.850} & \textbf{0.948} & \textbf{0.683} \\
FFT-16            & 16 & 0.843 & 0.942 & 0.658 \\
DWT-Haar-L4       & 12 & 0.841 & 0.938 & 0.665 \\
SWT-db4-L1        &  3 & 0.841 & 0.938 & 0.675 \\
SWT-db4-L2        &  6 & 0.840 & 0.941 & 0.659 \\
DWT-db4-L2        &  6 & 0.840 & 0.939 & 0.667 \\
FFT-8             &  8 & 0.840 & 0.937 & 0.671 \\
DWT-db4-L1        &  3 & 0.839 & 0.938 & 0.673 \\
DWT-db4-L4        & 12 & 0.838 & 0.938 & 0.644 \\
SWT-db4-L3        &  9 & 0.835 & 0.937 & 0.650 \\
FFT-4             &  4 & 0.831 & 0.929 & 0.661 \\
\bottomrule
\end{tabular}
\end{table}

\cref{tab:supp_freq} shows that DCT attains the highest in-domain values on BOCCHI across all three metrics (mIoU, Dice, BdF1), leading the second-best alternative by $0.70$, $0.53$, and $0.78$ percentage points respectively. The broader observation is that the benefit of the frequency branch does not come merely from choosing a particular handcrafted descriptor, but from explicitly providing frequency-domain evidence to the segmentation network: every frequency-domain variant substantially outperforms the spatial-only baseline reported in the main-paper architecture ablation (\emph{w/o Freq Branch}, $0.804$ in-domain mIoU on BOCCHI), and even the weakest alternative (FFT-4, $0.831$) clears that baseline by $2.7$ pp. We adopt DCT as our default because it produces real-valued, energy-concentrating coefficients, preserves explicit spatial locality from block-wise computation, and aligns naturally with the classical HiFST prior for blur detection~\cite{golestaneh2017hifst}.

\section{Architecture $\times$ Loss Interaction}
\label{sec:supp_arch_loss}

The main paper ablates architecture and loss independently with BCE / Dice / BCE+Dice. Here we revisit the BCE+Dice default against a natural loss-side alternative, Focal Frequency Loss~\cite{jiang2021focal} augmentation (BCE+Dice+Freq), to test whether an explicit frequency loss could substitute for our DCT branch. \cref{tab:supp_arch_loss} reports the joint ablation: each architecture variant trained under each of these two loss configurations. The trend is clear: removing the frequency branch causes the largest in-domain drop, while adding Focal Frequency Loss on top of BCE+Dice gives marginal or negative returns once the frequency branch is present. This is consistent with our interpretation that the explicit DCT branch already provides the frequency evidence that Focal Frequency Loss would otherwise supply through the loss function.

\begin{table}[t]
\centering
\small
\caption{\textbf{Architecture $\times$ loss interaction (in-domain on BOCCHI).} Each row is one combination of architecture variant and training loss.}
\label{tab:supp_arch_loss}
\begin{tabular}{l l c c}
\toprule
Architecture & Loss & Params (M) & In-domain mIoU$\uparrow$ \\
\midrule
Full MSDCT-UNet    & BCE+Dice        & 29.45 & \textbf{0.850} \\
Full MSDCT-UNet    & BCE+Dice+Freq   & 29.45 & 0.843 \\
w/o Freq Attn      & BCE+Dice        & 29.38 & 0.845 \\
w/o Freq Attn      & BCE+Dice+Freq   & 29.38 & 0.846 \\
w/o Freq Branch    & BCE+Dice        & 19.50 & 0.804 \\
w/o Freq Branch    & BCE+Dice+Freq   & 19.50 & 0.803 \\
\bottomrule
\end{tabular}
\end{table}

\section{Computational Efficiency}
\label{sec:supp_efficiency}

MSDCT-UNet prioritizes detection accuracy over inference speed. \cref{tab:supp_efficiency} reports parameter counts and inference throughput (FPS at $1080 \times 720$) for MSDCT-UNet and all 12 baselines, measured on the same RTX 5070 Ti GPU under identical conditions. Across the table, the evaluated methods span roughly three orders of magnitude in parameter count (from $0.11$M for ESMDL-UNet to $67.9$M for NAFNet) and two orders of magnitude in throughput (from $1.0$ to $179.3$ FPS), indicating a broad range of accuracy--efficiency operating points. MSDCT-UNet achieves the highest in-domain mIoU on BOCCHI but runs at only $1.0$ FPS, which is the main practical limitation of the current design. The bottleneck stems from the block-wise DCT preprocessing (which is CPU-heavy and not fully GPU-batched in our current implementation) and from the multi-scale DCT Attention computed at every encoder/decoder stage. Accelerating the DCT computation, for example via GPU-batched implementations or approximate DCT, is an important direction for practical deployment.

\begin{table}[t]
\centering
\small
\caption{\textbf{Parameters vs.\ inference speed vs.\ in-domain accuracy.} FPS measured at $1080 \times 720$ resolution on an NVIDIA RTX 5070 Ti (batch = 1, 20 warm-up + 100 measured iterations with \texttt{torch.cuda.synchronize()}, median reported). Models listed in ascending order of parameter count.}
\label{tab:supp_efficiency}
\begin{tabular}{l r r c}
\toprule
Model & Params (M) & FPS$\uparrow$ & BOCCHI mIoU$\uparrow$ \\
\midrule
ESMDL-UNet~\cite{li2025esmdlunet}         &  0.11 &  40.7 & 0.677 \\
MSDSeg~\cite{msdsegnet}                   &  2.30 &  66.5 & 0.714 \\
BiSeNetV2~\cite{yu2021bisenetv2}          &  2.42 & 125.5 & 0.759 \\
Cellpose3~\cite{stringer2021cellpose}     &  6.60 &  16.3 & 0.838 \\
STDC1~\cite{fan2021stdc}                  &  7.79 & 179.3 & 0.778 \\
STDC2~\cite{fan2021stdc}                  & 11.82 & 134.4 & 0.799 \\
DDRNet-23~\cite{hong2022ddrnet}           & 20.15 & 120.1 & 0.701 \\
MSDCT-UNet (Ours)                             & 29.45 &   1.0 & \textbf{0.850} \\
MSDU-Net~\cite{msdunet}                   & 33.79 &   8.2 & 0.796 \\
KDSNet-R50~\cite{kdsnet}                  & 39.76 &  38.2 & 0.783 \\
BEVANet~\cite{zhao2025bevanet}            & 58.60 &  96.8 & 0.778 \\
KDSNet-R101~\cite{kdsnet}                 & 58.75 &  29.3 & 0.773 \\
NAFNet~\cite{chen2022nafnet}              & 67.89 &   3.8 & 0.760 \\
\bottomrule
\end{tabular}
\end{table}

\section{Qualitative Results}
\label{sec:supp_qualitative}

\begin{figure}[t]
    \centering
    \makebox[\linewidth][c]{\includegraphics[width=1.4\linewidth]{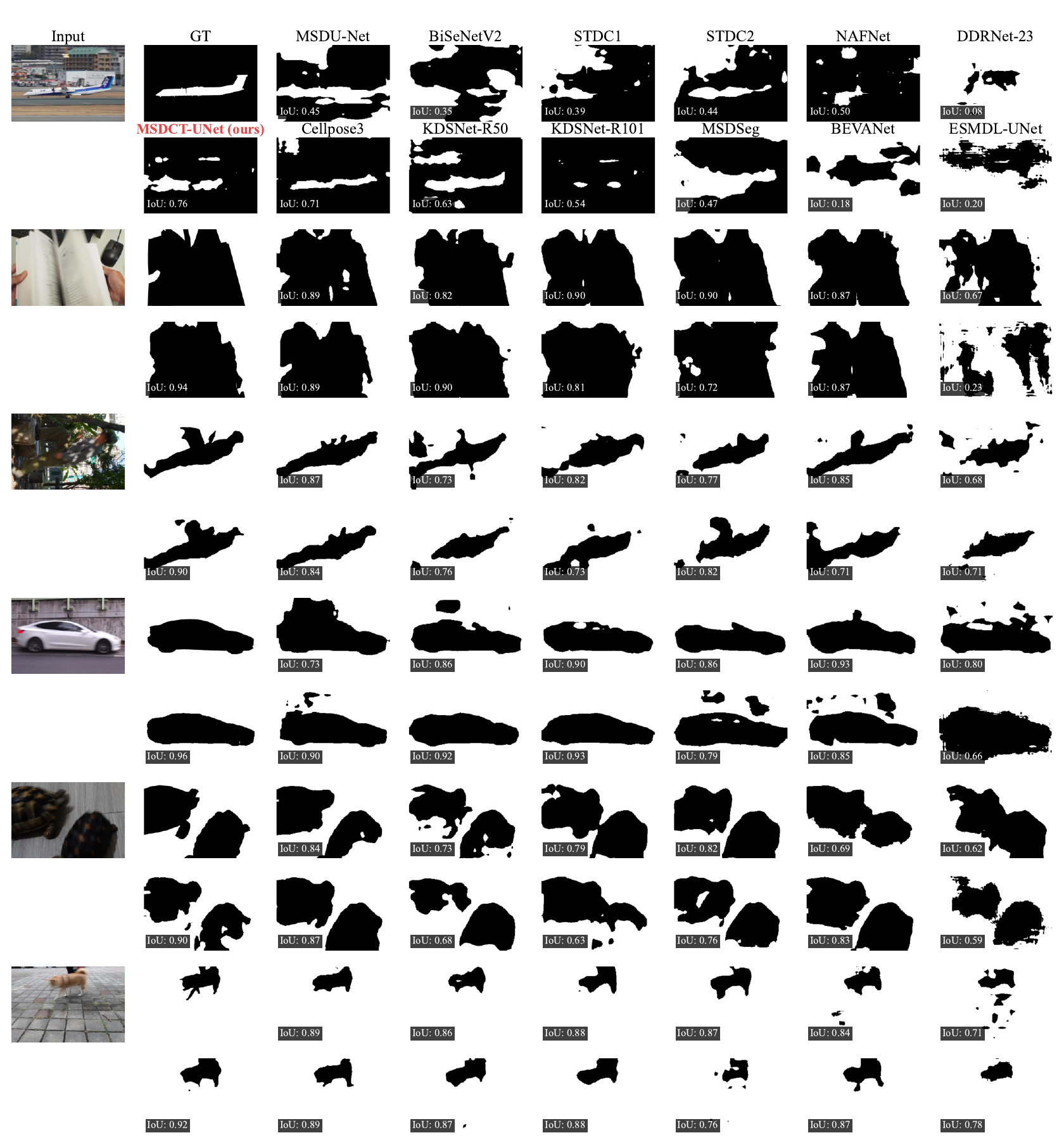}}
    \caption{\textbf{Full-baseline qualitative comparison on BOCCHI validation (success cases).} Each case occupies two rows: the top row shows Input, GT, and the six older baselines (MSDU-Net, BiSeNetV2, STDC1, STDC2, NAFNet, DDRNet-23); the bottom row shows MSDCT-UNet (ours) and the six newer baselines (Cellpose3, KDSNet-R50, KDSNet-R101, MSDSeg, BEVANet, ESMDL-UNet). Six success cases are stacked vertically. All predictions are binary masks (black = blur, white = sharp).}
    \label{fig:supp_grid_success}
\end{figure}

\begin{figure}[t]
    \centering
    \includegraphics[width=\linewidth]{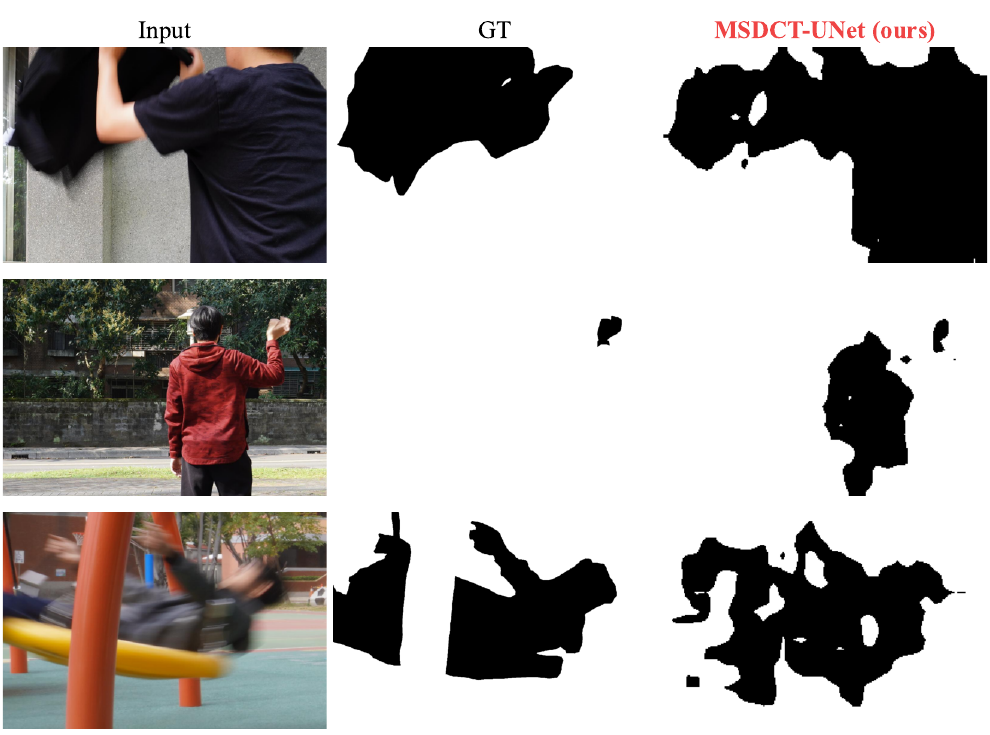}
    \caption{\textbf{Failure cases on BOCCHI validation (MSDCT-UNet only).} Three cases where MSDCT-UNet underperforms; columns are Input, GT, and MSDCT-UNet's prediction (binary mask).}
    \label{fig:supp_grid_failure}
\end{figure}

We provide full qualitative comparisons against all 12 paper baselines (\cref{fig:supp_grid_success}) and ours-only visualizations on failure cases (\cref{fig:supp_grid_failure}). Predictions are rendered as binary masks (white = predicted sharp, black = blur) for direct shape comparison.

\section{Cross-Dataset Per-class Analysis}
\label{sec:supp_cross}

\cref{tab:supp_cross_all} reports the full set of six cross-dataset metrics (mIoU, Dice, Precision, Recall, BdF1, BdIoU) for each of the thirteen evaluated models under each of the four training sources; all evaluations are performed on the Inference Dataset (572 images, per-image-mean aggregation) without fine-tuning. The dataset-level pattern observed in the main paper holds across the region-detection metrics: averaged over the thirteen models, BOCCHI-trained ones surpass ReLoBlur-trained ones on every metric (mIoU $0.563$ vs.\ $0.425$, $+13.8$ pp; Dice $0.814$ vs.\ $0.646$, $+16.8$ pp; Precision $0.808$ vs.\ $0.731$, $+7.7$ pp; Recall $0.875$ vs.\ $0.636$, $+23.9$ pp; BdF1 $0.264$ vs.\ $0.236$, $+2.8$ pp; BdIoU $0.152$ vs.\ $0.131$, $+2.1$ pp), and surpass CUHK$_{\text{motion}}$-trained ones by a wider margin on every metric. BOCCHI also outperforms OMoBlur (12-model Average, NAFNet $\times$ OMoBlur omitted) on the primary region-detection metrics: mIoU $+2.2$ pp, Dice $+7.9$ pp, Recall $+15.8$ pp. OMoBlur compensates with higher Precision ($+3.0$ pp) and boundary quality (BdF1 $+5.0$ pp, BdIoU $+6.9$ pp), reflecting its conservative synthetic-blur training. The dataset-level region-detection advantage holds independently of architecture: all $13$ of $13$ models follow BOCCHI $\geq$ ReLoBlur on mIoU.

\begin{table}[t]
\centering
\scriptsize
\setlength{\tabcolsep}{4pt}
\caption{\textbf{Full cross-dataset metrics per baseline and training source.} All thirteen models are evaluated on the Inference Dataset (572 images, per-image-mean aggregation) without fine-tuning. Each block corresponds to one training source. \textbf{Bold} marks the best value per column within a block. Prec.\ and Rec.\ are abbreviations for Precision and Recall.}
\label{tab:supp_cross_all}
\begin{tabular}{l c c c c c c}
\toprule
Model & mIoU$\uparrow$ & Dice$\uparrow$ & Prec.$\uparrow$ & Rec.$\uparrow$ & BdF1$\uparrow$ & BdIoU$\uparrow$ \\
\midrule
\multicolumn{7}{l}{\textit{Trained on BOCCHI}} \\
\midrule
MSDU-Net~\cite{msdunet}                       & 0.609 & 0.830 & 0.830 & 0.876 & 0.326 & 0.188 \\
BiSeNetV2~\cite{yu2021bisenetv2}              & 0.539 & 0.791 & 0.800 & 0.843 & 0.256 & 0.146 \\
STDC1~\cite{fan2021stdc}                      & 0.541 & 0.789 & 0.805 & 0.840 & 0.242 & 0.138 \\
STDC2~\cite{fan2021stdc}                      & 0.577 & 0.823 & 0.814 & 0.882 & 0.240 & 0.138 \\
NAFNet~\cite{chen2022nafnet}                  & 0.568 & 0.818 & 0.802 & 0.882 & 0.268 & 0.154 \\
DDRNet-23~\cite{hong2022ddrnet}               & 0.517 & 0.800 & 0.781 & 0.877 & 0.206 & 0.115 \\
Cellpose3~\cite{stringer2021cellpose}         & \textbf{0.662} & \textbf{0.858} & \textbf{0.841} & 0.910 & \textbf{0.423} & \textbf{0.255} \\
KDSNet-R50~\cite{kdsnet}                      & 0.557 & 0.809 & 0.813 & 0.871 & 0.231 & 0.130 \\
KDSNet-R101~\cite{kdsnet}                     & 0.557 & 0.794 & 0.815 & 0.835 & 0.232 & 0.131 \\
MSDSeg~\cite{msdsegnet}                       & 0.518 & 0.816 & 0.787 & 0.901 & 0.182 & 0.101 \\
BEVANet~\cite{zhao2025bevanet}                & 0.552 & 0.804 & 0.800 & 0.861 & 0.272 & 0.157 \\
ESMDL-UNet~\cite{li2025esmdlunet}             & 0.518 & 0.806 & 0.787 & 0.873 & 0.208 & 0.108 \\
MSDCT-UNet (Ours)                             & 0.610 & 0.848 & 0.823 & \textbf{0.925} & 0.350 & 0.211 \\
\cmidrule(lr){1-7}
\textit{Average}                              & \textit{0.563} & \textit{0.814} & \textit{0.808} & \textit{0.875} & \textit{0.264} & \textit{0.152} \\
\midrule
\multicolumn{7}{l}{\textit{Trained on ReLoBlur}} \\
\midrule
MSDU-Net~\cite{msdunet}                       & 0.556 & 0.758 & 0.817 & 0.736 & 0.315 & 0.173 \\
BiSeNetV2~\cite{yu2021bisenetv2}              & 0.470 & 0.695 & 0.801 & 0.651 & 0.216 & 0.119 \\
STDC1~\cite{fan2021stdc}                      & 0.283 & 0.518 & 0.615 & 0.538 & 0.202 & 0.112 \\
STDC2~\cite{fan2021stdc}                      & 0.269 & 0.443 & 0.579 & 0.437 & 0.230 & 0.130 \\
NAFNet~\cite{chen2022nafnet}                  & 0.478 & 0.730 & 0.792 & 0.718 & 0.227 & 0.127 \\
DDRNet-23~\cite{hong2022ddrnet}               & 0.462 & 0.704 & 0.791 & 0.665 & 0.187 & 0.103 \\
Cellpose3~\cite{stringer2021cellpose}         & 0.555 & 0.758 & \textbf{0.822} & 0.744 & 0.285 & 0.160 \\
KDSNet-R50~\cite{kdsnet}                      & 0.275 & 0.465 & 0.594 & 0.465 & 0.236 & 0.130 \\
KDSNet-R101~\cite{kdsnet}                     & 0.274 & 0.433 & 0.574 & 0.423 & 0.256 & 0.142 \\
MSDSeg~\cite{msdsegnet}                       & 0.463 & 0.726 & 0.776 & 0.711 & 0.196 & 0.108 \\
BEVANet~\cite{zhao2025bevanet}                & 0.484 & 0.718 & 0.796 & 0.692 & 0.226 & 0.125 \\
ESMDL-UNet~\cite{li2025esmdlunet}             & 0.391 & 0.631 & 0.759 & 0.593 & 0.157 & 0.084 \\
MSDCT-UNet (Ours)                             & \textbf{0.563} & \textbf{0.815} & 0.795 & \textbf{0.897} & \textbf{0.335} & \textbf{0.193} \\
\cmidrule(lr){1-7}
\textit{Average}                              & \textit{0.425} & \textit{0.646} & \textit{0.731} & \textit{0.636} & \textit{0.236} & \textit{0.131} \\
\midrule
\multicolumn{7}{l}{\textit{Trained on CUHK$_{\text{motion}}$}} \\
\midrule
MSDU-Net~\cite{msdunet}                       & 0.142 & 0.307 & 0.525 & 0.268 & 0.204 & 0.114 \\
BiSeNetV2~\cite{yu2021bisenetv2}              & 0.248 & 0.445 & 0.587 & 0.455 & 0.062 & 0.034 \\
STDC1~\cite{fan2021stdc}                      & 0.197 & 0.310 & 0.585 & 0.286 & 0.101 & 0.055 \\
STDC2~\cite{fan2021stdc}                      & 0.182 & 0.288 & 0.478 & 0.277 & 0.085 & 0.046 \\
NAFNet~\cite{chen2022nafnet}                  & 0.285 & 0.529 & 0.688 & \textbf{0.533} & 0.113 & 0.060 \\
DDRNet-23~\cite{hong2022ddrnet}               & \textbf{0.292} & \textbf{0.548} & \textbf{0.715} & 0.490 & 0.116 & 0.063 \\
Cellpose3~\cite{stringer2021cellpose}         & 0.160 & 0.324 & 0.501 & 0.306 & \textbf{0.217} & \textbf{0.123} \\
KDSNet-R50~\cite{kdsnet}                      & 0.225 & 0.424 & 0.660 & 0.369 & 0.121 & 0.067 \\
KDSNet-R101~\cite{kdsnet}                     & 0.230 & 0.432 & 0.658 & 0.372 & 0.132 & 0.071 \\
MSDSeg~\cite{msdsegnet}                       & 0.141 & 0.128 & 0.319 & 0.121 & 0.060 & 0.033 \\
BEVANet~\cite{zhao2025bevanet}                & 0.271 & 0.493 & 0.701 & 0.434 & 0.114 & 0.062 \\
ESMDL-UNet~\cite{li2025esmdlunet}             & 0.231 & 0.397 & 0.680 & 0.352 & 0.139 & 0.071 \\
MSDCT-UNet (Ours)                             & 0.183 & 0.290 & 0.563 & 0.226 & 0.162 & 0.089 \\
\cmidrule(lr){1-7}
\textit{Average}                              & \textit{0.214} & \textit{0.378} & \textit{0.589} & \textit{0.345} & \textit{0.125} & \textit{0.068} \\
\midrule
\multicolumn{7}{l}{\textit{Trained on OMoBlur (NAFNet $\times$ OMoBlur omitted, $n=12$)}} \\
\midrule
MSDU-Net~\cite{msdunet}                       & \textbf{0.616} & 0.805 & 0.854 & 0.810 & 0.359 & 0.263 \\
BiSeNetV2~\cite{yu2021bisenetv2}              & 0.526 & 0.704 & 0.843 & 0.666 & 0.311 & 0.222 \\
STDC1~\cite{fan2021stdc}                      & 0.496 & 0.657 & 0.839 & 0.597 & 0.296 & 0.217 \\
STDC2~\cite{fan2021stdc}                      & 0.530 & 0.705 & 0.854 & 0.655 & 0.310 & 0.225 \\
DDRNet-23~\cite{hong2022ddrnet}               & 0.489 & 0.673 & 0.827 & 0.626 & 0.297 & 0.172 \\
Cellpose3~\cite{stringer2021cellpose}         & 0.628 & \textbf{0.825} & \textbf{0.859} & \textbf{0.848} & 0.357 & 0.264 \\
KDSNet-R50~\cite{kdsnet}                      & 0.528 & 0.717 & 0.843 & 0.684 & 0.314 & 0.228 \\
KDSNet-R101~\cite{kdsnet}                     & 0.525 & 0.718 & 0.847 & 0.696 & 0.298 & 0.220 \\
MSDSeg~\cite{msdsegnet}                       & 0.547 & 0.788 & 0.811 & 0.828 & 0.304 & 0.219 \\
BEVANet~\cite{zhao2025bevanet}                & 0.535 & 0.746 & 0.837 & 0.731 & 0.310 & 0.182 \\
ESMDL-UNet~\cite{li2025esmdlunet}             & 0.458 & 0.698 & 0.788 & 0.684 & 0.256 & 0.180 \\
MSDCT-UNet (Ours)                             & 0.610 & 0.783 & 0.854 & 0.781 & \textbf{0.359} & \textbf{0.266} \\
\cmidrule(lr){1-7}
\textit{Average} ($n{=}12$)                   & \textit{0.541} & \textit{0.735} & \textit{0.838} & \textit{0.717} & \textit{0.314} & \textit{0.221} \\
\bottomrule
\end{tabular}
\end{table}

\section{Cross-Dataset Qualitative Comparison}
\label{sec:supp_cross_qual}

\begin{figure}[t]
    \centering
    \includegraphics[width=\linewidth]{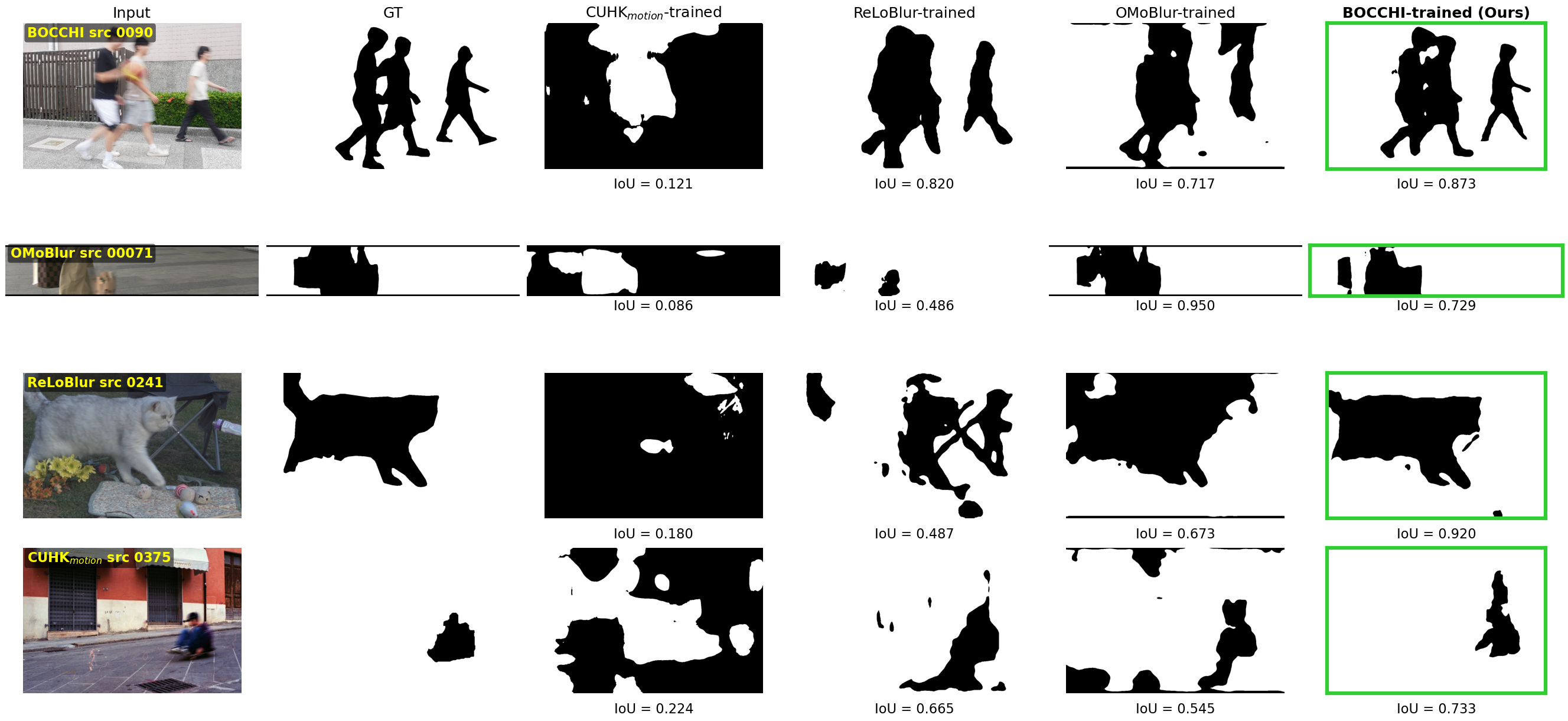}
    \caption{\textbf{Cross-dataset qualitative comparison.} The same MSDCT-UNet architecture trained from scratch on four sources (CUHK$_{\text{motion}}$, ReLoBlur, OMoBlur, BOCCHI) and evaluated on four Inference Dataset images (no fine-tuning). Per-cell numbers are IoU; \textbf{BOCCHI-trained (green box) wins all four cases} ($0.873$, $0.729$, $0.920$, $0.733$). Masks: white $=$ predicted sharp / black $=$ predicted blur.}
    \label{fig:supp_cross_qual}
\end{figure}

To visualise the dataset-level advantage quantified in \cref{tab:supp_cross_all}, \cref{fig:supp_cross_qual} shows the same MSDCT-UNet architecture trained from scratch on each of the four sources (CUHK$_{\text{motion}}$, ReLoBlur, OMoBlur, BOCCHI) and evaluated on four representative images sampled from the held-out Inference Dataset (no fine-tuning). Cases are drawn from four distinct source-subsets of the benchmark (row~1 from the BOCCHI protocol subset, row~2 from the OMoBlur subset, row~3 from the ReLoBlur subset, row~4 from the CUHK$_{\text{motion}}$ subset) to avoid cherry-picking any single subset.

Across the four cases, the BOCCHI-trained MSDCT-UNet recovers the most faithful boundary; the CUHK$_{\text{motion}}$-trained variant over-segments background regions, the ReLoBlur-trained variant leaves the foreground partially unmarked, and the OMoBlur-trained variant produces fragmented or over-eager predictions. This visual pattern directly mirrors the quantitative dataset-level dominance reported in \cref{tab:supp_cross_all}: BOCCHI is the strongest training source on every primary metric, and the advantage holds across all four source-subsets of the Inference Dataset rather than only on the BOCCHI-protocol subset.

\section{Cross-Dataset Transfer: Domain Gap Analysis}
\label{sec:supp_domain_gap}

The in-domain to cross-dataset performance gap observed across all evaluated methods (e.g., BOCCHI in-domain mIoU $0.850$ vs.\ BOCCHI-trained cross-dataset mIoU $0.610$ for MSDCT-UNet) deserves a dedicated discussion, since it is a property of the benchmark setup rather than of any single model.

\textbf{On BOCCHI's high in-domain mIoU.} BOCCHI's in-domain mIoU of $0.850$ may appear high relative to its claim of being a more challenging benchmark. This number reflects the low average blur ratio ($0.22$) which biases mIoU toward the dominant sharp class; BOCCHI's hardness instead manifests in its resistance to gradient-based shortcuts: a frequency-agnostic spatial backbone (DDRNet-23) trails MSDCT-UNet by $2.0$ pp on ReLoBlur but by $14.9$ pp on BOCCHI ($7.5\times$ amplification), confirming that BOCCHI penalizes spatial-only shortcuts more strongly than existing benchmarks.

\textbf{Three axes of heterogeneity.} The four training sources (BOCCHI, OMoBlur, ReLoBlur, CUHK$_{\text{motion}}$) differ along at least three axes that jointly contribute to the observed gap:
(i) \emph{Annotation granularity.} ReLoBlur and BOCCHI annotations both concentrate on small-area blur regions (average blur ratio $0.20$ and $0.22$), but BOCCHI pairs them with bimodal sharp surroundings that overlap the blur gradient distribution; CUHK$_{\text{motion}}$ has wider blur-ratio coverage but only 200 samples; OMoBlur~\cite{yu2026omoblur} uses optical-flow-derived masks as soft supervision for restoration rather than pixel-precise detection labels. A model trained on one annotation regime is not immediately calibrated for another.
(ii) \emph{Camera point-spread function (PSF).} Each dataset is captured with a different camera population (BOCCHI: five consumer cameras; ReLoBlur: predominantly street-camera setup; CUHK$_{\text{motion}}$: heterogeneous consumer imagery; OMoBlur: a programmable industrial camera with overlapping-exposure synthesis). The resulting PSF distributions differ in lens blur, sensor noise, and motion-vector characteristics, which are not explicitly modelled by any evaluated method.
(iii) \emph{Blur-ratio regime.} As shown in Fig.~3 of the main paper, the blur-ratio distributions span varying ranges (ReLoBlur and BOCCHI concentrate at low ratios, OMoBlur at moderate ratios, CUHK$_{\text{motion}}$ broadly distributed); cross-source training therefore requires substantial distribution shift that current methods do not explicitly account for.

\textbf{Why simple remedies are insufficient.} Two natural remedies, namely unified cross-dataset training and na\"ive fine-tuning, each have drawbacks in this setting. Unified training tends to overfit to the largest source (ReLoBlur, $1{,}200$ samples) and dilute the signal of the harder regime that BOCCHI exposes. Na\"ive fine-tuning on a new source typically induces catastrophic forgetting of the source-dataset accuracy, a well-known issue in continual learning. The large source-specific in-domain headroom observed in our experiments ($\geq 0.85$ on BOCCHI, $\geq 0.93$ on ReLoBlur for MSDCT-UNet) suggests that continual or multi-source strategies that preserve prior-dataset performance are a promising direction, and we believe this is a substantial open problem for local blur detection under heterogeneous data sources.

\FloatBarrier